\documentclass[preprint,5p,times,twocolumn]{elsarticle}

\usepackage{hyperref}

\biboptions{sort&compress}

\usepackage{color}
\usepackage{tabularx} 
\usepackage{bigstrut}
\usepackage{longtable} 
\usepackage{adjustbox}
\usepackage{booktabs}
\usepackage{amsthm}
\usepackage{amssymb}
\usepackage{comment}

\theoremstyle{definition}
\newtheorem{definition}{Definition}[section]
\usepackage{amsmath,amsfonts,amsthm}
\usepackage{upgreek}
\usepackage{algorithmic}
\usepackage[ruled]{algorithm2e} 
\usepackage{multirow}
\usepackage{colortbl}
\usepackage{rotating}

\newif\ifproofread
\newcommand{\change}[1]{%
	\ifproofread
	\textcolor{red}{#1}%
	\else
	#1%
	\fi
}

\newif\iftkcommnets
\newcommand{\tkcomm}[1]{%
	\iftkcommnets
	\textcolor{blue}{#1}%
	\else
	#1%
	\fi
}

\newcommand{\myrowcolour}{\rowcolor[gray]{0.925}}

\begin{document}

\begin{frontmatter}

\title{\change{A survey on Machine Learning-based Performance Improvement of Wireless Networks: PHY, MAC and Network layer}}

\author[1]{Merima Kulin \corref{correspondingauthor}}
\cortext[correspondingauthor]{Corresponding author.}
\ead{merima.kulin@ugent.be}

\author[2]{Tarik Kazaz}
\author[1]{Ingrid Moerman}
\author[1]{Eli de Poorter}



\address[1]{Ghent University, Department of Information Technology, B-9052 Gent, Belgium}
\address[2]{Delft University of Technology, Faculty of EEMCS, 2628 CD Delft, The Netherlands}

\begin{abstract}
\change{This paper provides a systematic and comprehensive survey that reviews the latest research efforts focused on machine learning (ML) based performance improvement of wireless networks, while considering all layers of the protocol stack (PHY, MAC and network). First, the related work and paper contributions are discussed, followed by providing the necessary background \tkcomm{on data-driven approaches and machine learning for non-machine learning experts} to understand all discussed techniques. Then, a comprehensive review is presented on works employing ML-based approaches to optimize the wireless communication parameters \tkcomm{settings} to achieve improved network quality-of-service (QoS) and quality-of-experience (QoE). We first categorize these works into: radio analysis, MAC analysis and network prediction approaches, followed by subcategories within each.} Finally, open challenges and broader perspectives are discussed.

\end{abstract}

\begin{keyword}
Machine learning, data science, deep learning, cognitive radio networks, protocol layers, MAC, PHY, performance optimization
\end{keyword}

\end{frontmatter}

\section{Introduction}
\label{sec:intro}
\noindent Science and the way we undertake research is rapidly changing.
The increase of data generation is present in all scientific disciplines \cite{abbasi2016big}, such as computer vision, speech recognition, finance (risk analytics), marketing and sales (e.g. customer churn analysis), pharmacy (e.g. drug discovery), personalized health-care (e.g. biomarker identification in cancer research), precision agriculture (e.g. crop lines detection, weeds detection...), politics (e.g. election campaigning), etc. 
Until the recent years, this trend has been less pronounced in the wireless networking domain, mainly due to the lack of ‘big data’ and 'big' communication capacity \cite{qian2017survey}. However, with the era of the Fifth Generation (5G) cellular systems and the Internet-of-Things (IoT), the big data deluge in the wireless networking domain is under way. For instance, massive amounts of data are generated by the omnipresent sensors used in smart cities \cite{rathore2016iot, nguyen2016traffic} (e.g. to monitor parking spaces availability in the cities, or monitor the conditions of  road traffic to manage and control traffic flows), smart infrastructures (e.g. to monitor the condition of railways or bridges), precision farming \cite{lottes2017uav, sa2018weednet} (e.g. monitor yield status,  soil temperature and humidity), environmental monitoring (e.g. pollution, temperature, precipitation sensing), 
IoT smart grid networks \cite{strohbach2015towards} (e.g. to monitor distribution grids or track energy consumption for demand forecasting), etc. 
\change{It is expected that 28.5 billion devices will be connected by 2022
	to the Internet \cite{cisco2019}}, which will create a huge global network of “things” and the demand for wireless resources will accordingly increase in an unprecedented way. On the other hand, the  set of available communication technologies is expanding (e.g. the release of the new IEEE 802.11 standards such as IEEE 802.11ax and IEEE 802.11ay; and 5G technologies), which compete for the same finite and limited radio spectrum resources pressuring the need for enhancing their coexistence and more effective use the scarce spectrum resources.
Similarly, on the mobile systems landscape, mobile data usage is tremendously increasing; according to the latest Ericsson’s mobility report there are now 5.9 billion mobile broadband subscriptions globally, generating more than 25 exabytes per month of wireless data traffic \cite{ericsonRep}, a growth close to $88\%$ between Q4 2017 and Q4 2018!

So, big data today is a reality! 
\begin{figure*}[t!]
	\centering
	\includegraphics[width=0.73\textwidth]{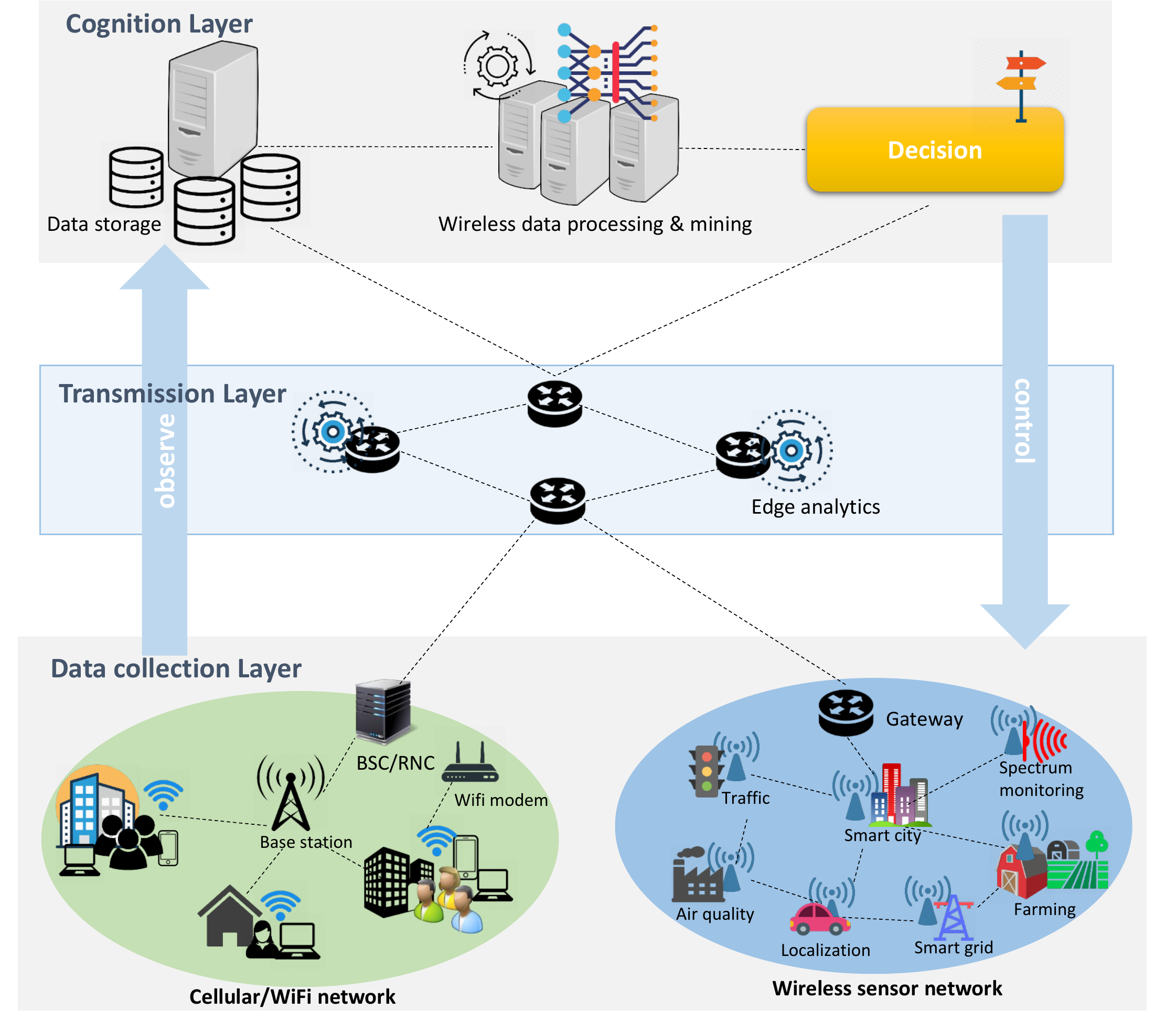}
	\caption{Architecture for wireless big data analysis}
	\label{fig:intro:wirelessBigData}
\end{figure*}

\change{However, wireless networks and the generated traffic patterns are becoming more and more complex and challenging to understand. 
	For instance, wireless networks yield many network \textit{performance indicators} (e.g. signal-to-noise ratio (SNR), link access success/collision
	rate, packet loss rate, bit error rate (BER), latency, link quality indicator, throughput, energy consumption, etc.) and \textit{operating parameters}  at different layers of the network protocol stack (e.g. at the PHY layer: frequency channel, modulation scheme, transmitter power; at the MAC layer:  MAC protocol selection, and parameters of specific MAC protocols such as CSMA: contention window size, maximum number of backoffs, backoff exponent; TSCH: channel hopping sequence, etc.) having significant impact on the communication performance.}

\change{Tuning of these operating parameters and achieving cross-layer optimization to maximize the end-to-end performance is a challenging task. This is especially complex due to the huge traffic demands and heterogeneity of deployed wireless technologies. 	
	To address these challenges, machine learning (ML) is increasingly used
	to develop  advanced approaches that can autonomously extract \textit{patterns} and \textit{predict} trends (e.g. at the PHY layer: interference recognition, at the MAC layer: link quality prediction, at the network layer: traffic demand estimation) based on environmental measurements and performance indicators as input. 
	Such patterns can be used to optimize the parameter settings at different protocol layers, e.g PHY, MAC or network layer.}

\change{For instance, consider Figure \ref{fig:intro:wirelessBigData}, which illustrates an architecture with heterogeneous wireless access technologies, capable of \textit{collecting} large amounts of observations from the wireless devices, \textit{processing} them and feeding into ML algorithms which generate patterns that can help making better decisions to optimize the operating parameters and improve the network quality-of-service (QoS) and quality-of-experience QoE.}

\change{Obviously, there is an urgent need for the development of novel intelligent solutions to improve the wireless networking performance.} 
This has motivated this paper and its main goal to raise awareness of the emerging interdisciplinary research area (spanning wireless networks and communications, machine learning, statistics, experimental-driven research and other research disciplines) and showcase the state-of-the-art on how to apply ML to improve the performance of wireless networks to solve the challenges that the wireless community is currently facing.

\change{Although several survey papers exist, most of them focus on ML in a specific domain or network layer. To the best of our knowledge, this is the first survey that comprehensively reviews the latest research efforts focused on ML-based performance improvements of wireless networks while considering all layers of the protocol stack (PHY, MAC and network), whilst also providing the necessary \tkcomm{tutorial} for non-machine learning experts to understand all discussed techniques.}

\tkcomm{
	\textbf{Paper organization}: We structure this paper as shown on Figure \ref{fig:paperOutline}.
	\begin{figure*}[th]
		\centering
		\includegraphics[width=1.0\textwidth]{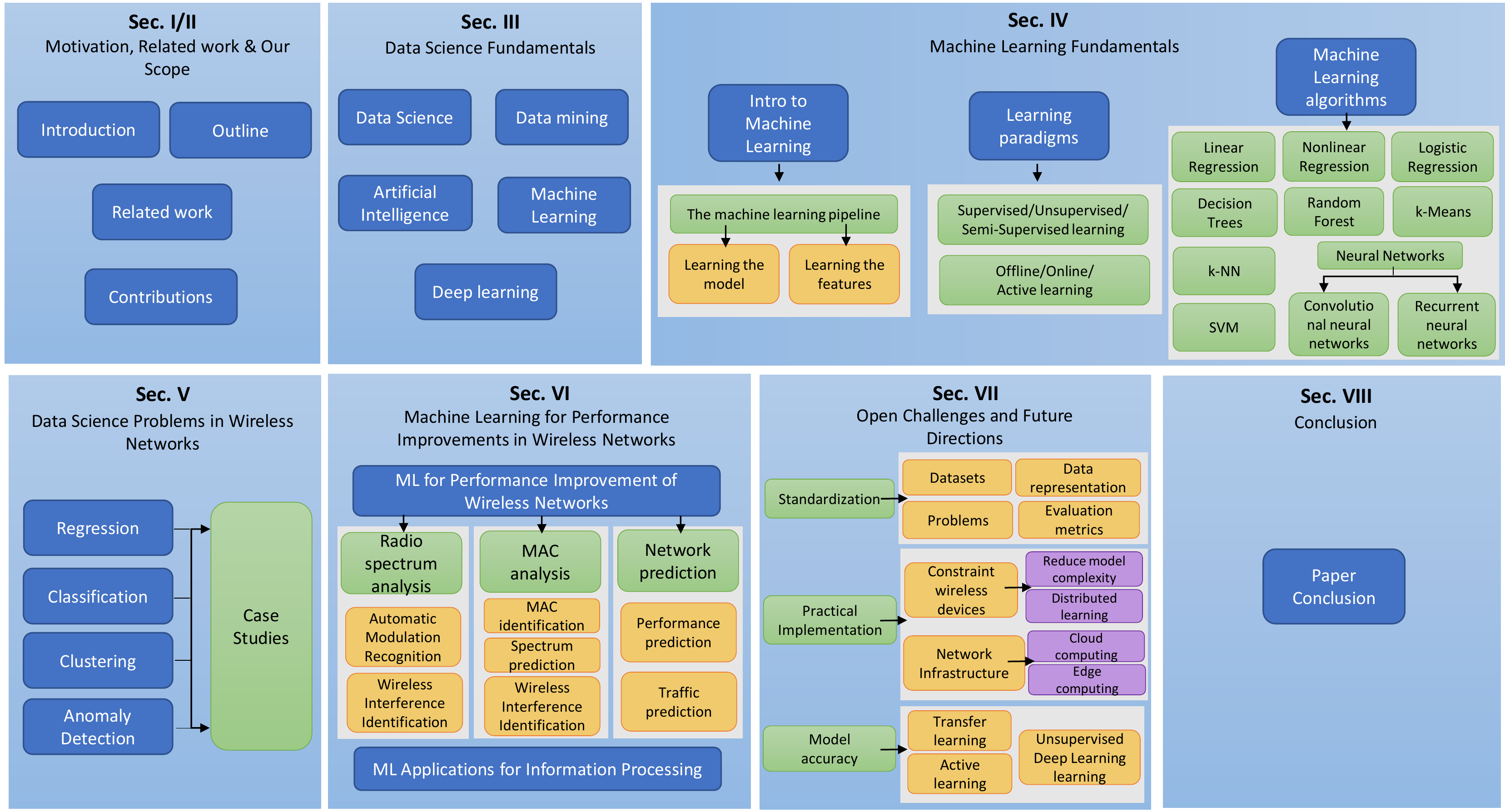}
		\caption{Paper outline}
		\label{fig:paperOutline}
	\end{figure*}
We start with discussing the related work and distinguishing our work with the state-of-the-art, in Section \ref{sec:2}. We conclude that section with a list of our contributions. In Section \ref{sec:3}, we present a high-level introduction to data science, data mining, artificial intelligence, machine learning and deep learning. The main goal here is to define these interchangeably used terms and how they related to each other. In \ref{sec:4} we provide a tutorial focused on machine learning, we overview various types of learning paradigms and introduce a couple of popular machine learning algorithms. Section \ref{subsec:learningProblems} introduces four common types of data-driven problems in the context of wireless networks and provides examples of several case studies. The objective of this section is to help the reader formulate a wireless networking problem into a data-driven problem suitable for machine learning. Section \ref{sub:mlsurvey} discusses the latest state-of-the-art about machine learning for performance improvements of wireless networks. First, we categorize these works into: radio analysis, MAC analysis and network prediction approaches; then we discuss example works within each category and give an overview in tabular form, looking at various aspects including: input data,  learning approach and algorithm, type of wireless network, achieved performance improvement, etc. In Section \ref{sec:challenges}, we discuss open challenges and present future directions for each. Section \ref{sec:concl} concludes the paper.}

\tkcomm{\section{Related Work and Our Contributions}}
\label{sec:2}
\subsection{Related Work}
With the advances in hardware and computing power and the ability to collect, store and process massive amounts of data, machine learning (ML) has found its way into many different scientific fields. The challenges faced by current 5G and future wireless networks pushed also the wireless networking domain to seek innovative solutions to ensure expected network performance. To address these challenges, ML is increasingly used in wireless networks. In parallel, a growing number of surveys and tutorials are emerging on ML for future wireless networks. \tkcomm{Table \ref{tab:intro:relatedwork} provides an overview and comparison with the existing survey papers.} For instance:

\begin{table*}[bt]
	\centering
	\begin{adjustbox}{width=1.0\textwidth}
		\centering
		\begin{tabular}{p{1cm} p{2.5cm} p{4cm} p{4cm} l p{1cm}}
			\toprule
			\textbf{Paper} & \textbf{Tutorial on ML} & \textbf{Wireless network} & \textbf{Application Area}& \textbf{ML paradigms} & \textbf{Year}\\
			\toprule
			\hline
			
			\cite{bkassiny2012survey} & \checkmark & CRN & Decision-making and feature classification in CRN & Supervised, unsupervised and reinforcement learning  & 2012\\
			\cite{alsheikh2014machine} & \checkmark &  localization, security, event detection, routing, data aggregation, MAC& WSN & Supervised, unsupervised and reinforcement learning & 2014\\
			\cite{wang2015artificial} & +- & HetNets & Self-configuration, self-healing, and self-optimization & AI-based techniques & 2015\\	
			\cite{ahad2016neural}& +- & CRN, WSN, Cellular and Mobile ad-hoc networks & Security, localization, routing, load balancing & NN & 2016\\
			\cite{park2016learning} &  & IoT & Big data analytics, event detection, data aggregation, etc.& Supervised, unsupervised and reinforcement learning & 2016 \\
			\cite{klaine2017survey} & \checkmark & Cellular networks & Self-configuration, self-healing, and self-optimization & Supervised, unsupervised and reinforcement learning& 2017\\
			\cite{zho2018intelligent} & +- & CRN & Spectrum sensing and access & Supervised, unsupervised and reinforcement learning& 2018\\
			\cite{mao2018deep} & +- & IoT, Cellular networks, WSN, CRN & Routing, resource allocation, security, signal detection, application identification, etc. & Deep learning &  2018 \\
			\cite{mohammadi2018deep}  & +-& IoT  & Big data and stream analytics & Deep learning  & 2018 \\
			\cite{chen2019artificial} & \checkmark & IoT, Mobile networks, CRN, UAV & Communication, virtual reality and edge caching & ANN & 2019 \\
			\cite{li2019survey} & +- & CRN & Signal Recognition & Deep learning & 2019 \\
			\cite{din2019machine} & +-  & IoT  & Smart cities &  Supervised, unsupervised and deep learning &  2019\\
			\cite{luong2019applications} & +- & Communications and networking & Wireless caching, data offloading, network security, traffic routing, resource sharing, etc. & Reinforcement learning & 2019 \\
			This & \checkmark & IoT, WSN, cellular networks, CRN &Performance improvement of wireless networks & Supervised, unsupervised and Deep learning  & 2019 \\
			\bottomrule	
		\end{tabular}
	\end{adjustbox}
	\caption{Overview of the related work}
	\label{tab:intro:relatedwork}
\end{table*}


\begin{itemize}
	\item In \cite{bkassiny2012survey}, the authors surveyed existing ML-based methods to address problems in Cognitive Radio Networks
	(CRNs).
	\item The authors of \cite{alsheikh2014machine} survey ML approaches in  WSNs (Wireless Sensor Networks) for various applications including location, security, routing, data aggregation and MAC.
	\item The authors of \cite{wang2015artificial} surveyed the state-of-the-art Artificial Intelligence (AI)-based techniques applied to heterogeneous networks (HetNets) focusing on the research issues of self-configuration, self-healing, and self-optimization.
	\item ML algorithms and their applications in self organizing cellular networks also focusing on 
	self-configuration, self-healing, and self-optimization, are surveyed in \cite{klaine2017survey}.
	\item In \cite{zho2018intelligent} ML applications in CRN are surveyed, that enable spectrum and energy efficient communications in dynamic wireless environments.
	\item  The authors of \cite{chen2019artificial} studied neural networks-based solutions to solve problems in wireless networks such as communication, virtual reality and edge caching.
	\item In \cite{ahad2016neural}, various applications of neural networks (NN) in wireless networks including security, localization, routing, load balancing are surveyed.
	\item The authors of \cite{park2016learning} surveyed ML techniques used in IoT networks for big data analytics, event detection, data aggregation, power control and other applications.
	\item Paper \cite{mao2018deep} surveys deep learning applications in wireless networks looking at aspects such as routing, resource allocation, security, signal detection, application identification, etc.
	\item Paper \cite{mohammadi2018deep} surveys deep learning applications in IoT networks for big data and stream analytics.
	\item Paper \cite{li2019survey} studies and surveys deep learning applications in cognitive radios for signal recognition tasks.
	\item The authors of \cite{din2019machine} survey ML approaches in the context of IoT smart cities.
	\item Paper \cite{luong2019applications} surveys reinforcement learning applications for various applications including network access and rate control, wireless caching, data offloading, network security, traffic routing, resource sharing, etc.
\end{itemize}

Nevertheless, some of the aforementioned works focus on reviewing specific wireless networking tasks (for example, wireless signal recognition \cite{li2019survey}), some focus on the application of specific ML techniques (for instance, deep learning \cite{ahad2016neural}, \cite{chen2019artificial}, \cite{li2019survey}) while some focus on the aspects of a specific wireless environment looking at broader 
applications (e.g. CRN \cite{bkassiny2012survey}, \cite{zho2018intelligent}, \cite{li2019survey}, and IoT \cite{park2016learning}, \cite{din2019machine}).
Furthermore, we noticed that some works miss out the necessary fundamentals for the readers who seek to learn the basics of an area outside their specialty. Finally, no existing work focuses on the literature on how to apply ML techniques to improve wireless network performance looking at possibilities at different layers of the network protocol stack.

To fill this gap, this paper provides a comprehensive introduction to ML for wireless networks and a 
survey of the latest advances in ML applications for performance improvement to address various challenges future wireless networks are facing.
We hope that this paper can help readers develop perspectives on and identify trends of this field and foster more subsequent studies on this topic.


\subsection{Contributions}

The main contributions of this paper are as follows:
\begin{itemize}
	\item \change{Introduction for non-machine learning experts to the necessary fundamentals}  on ML, AI, big data and data science in the context of wireless networks, \change{with numerous examples}. It examines when, why and how to use ML. 
	\item A systematic and comprehensive survey on the state-of-the-art that i) demonstrates the diversity of challenges impacting the  wireless networks performance that can be addressed with ML approaches and which ii) illustrates how ML is applied to improve the performance of wireless networks from various perspectives: PHY, MAC and the network layer.
	\item References to the latest research works (up to and including 2019) in the field of predictive ML approaches for improving the performance of wireless networks.
	\item Discussion on open challenges and future directions in the field.
\end{itemize}

\section{\change{Data Science Fundamentals}}
\label{sec:3}

\begin{figure}[bt]
	\centering
	\includegraphics[width=0.5\textwidth]{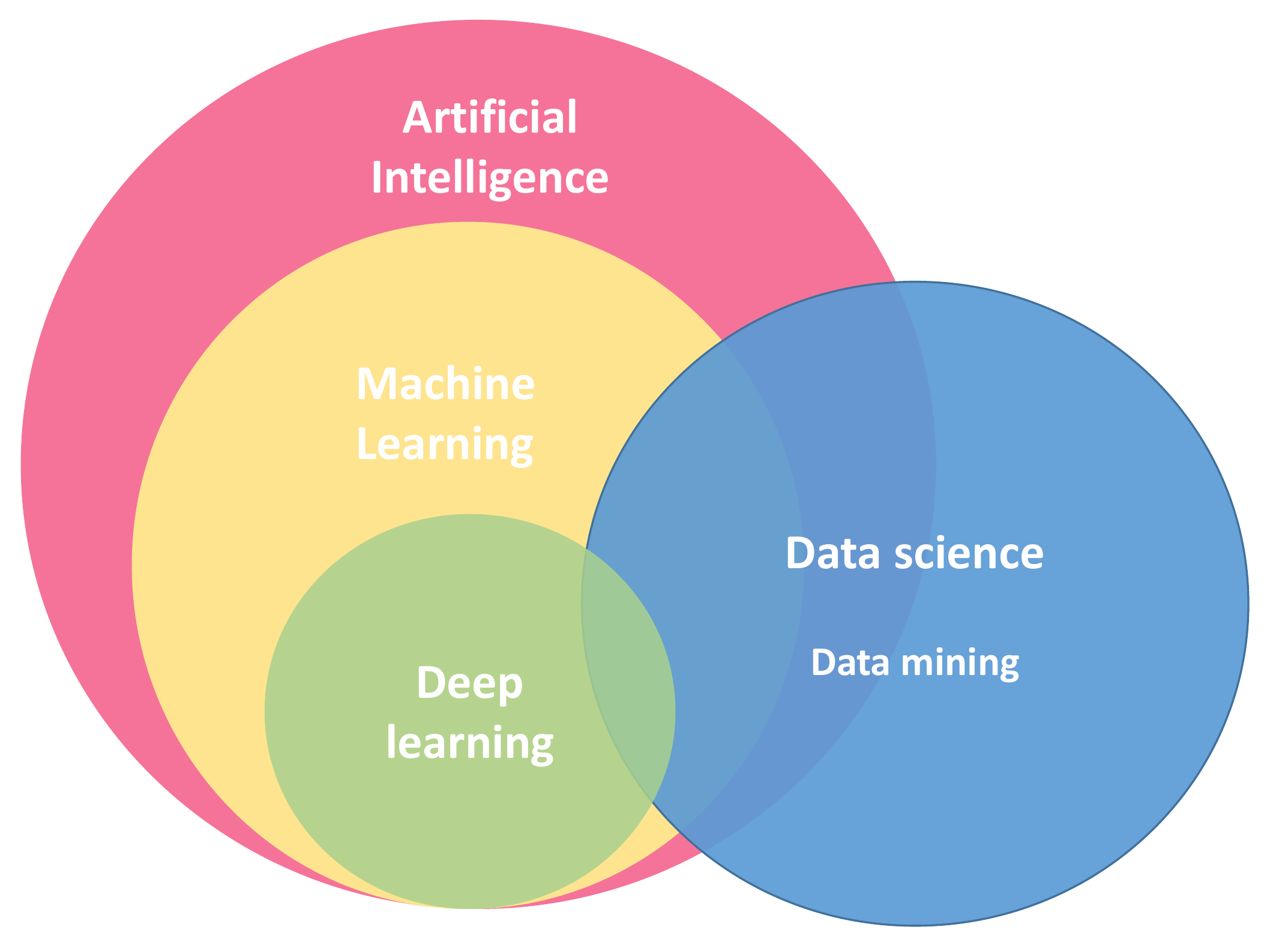}
	\caption{Data science vs. data mining vs. AI vs. ML vs. deep learning}
	\label{fig:intro:ml_dl}
\end{figure}

\noindent \tkcomm{The objective of this section is to introduce disciplines closely related to data-driven research and machine learning, and how they related to each other.}
Figure \ref{fig:intro:ml_dl} shows a Venn diagram, which illustrates the relation between data science, data mining, artificial intelligence (AI), machine learning and deep learning (DL), explained in more detail in the following subsections. This survey, particularly, focuses on ML/DL approaches in the context of wireless networks.

\subsection{Data Science}
\noindent\change{\textit{Data science} is the scientific discipline that studies everything related to data, from data acquisition, data storage, data analysis, data cleaning, data visualization, data interpretation, making decisions based on data, determining how to create value from data and how to communicate insights relevant to the business.
	One definition of the term data science, provided by Dhar \cite{dhar2013data}, is:
	\theoremstyle{definition}
	\begin{definition}
		\label{datascience_def}
		\textit{Data science is the study of the generalizable extraction of knowledge from data.} 
	\end{definition}
	Data science makes use of data mining, machine learning, AI techniques and also other approaches such as: heuristics algorithms, operational research, statistics, causal inference, etc. Practitioners of data science are typically skilled in mathematics, statistics, programming, machine learning, big data tools and communicating \tkcomm{the results}.}

\subsection{Data Mining}
\noindent\change{\textit{Data mining} aims to understand and discover new, previously unseen knowledge in the data. The term \textit{mining} refers to extracting content by digging. Applying this analogy to data, it may mean to extract insights by digging into data. A simple definition of data mining is:
	\theoremstyle{definition}
	\begin{definition}
		\label{dm_def}
		\textit{Data mining refers to the application of algorithms for extracting patterns from data.} 
	\end{definition}
	Compared to ML, data mining tends to focus on solving actual problems encountered in practice by exploiting algorithms developed by the ML community.
	For this purpose, a data-driven problem is first translated into a suitable data mining method \cite{kulin2016data}, which will be in detail discussed in Section \ref{subsec:learningProblems}.}

\subsection{Artificial Intelligence}

\noindent Artificial intelligence (AI) is concerned with making machines smart aiming to create a system which behaves like a human. This involves fields such as robotics, natural language processing, information retrieval, computer vision and \textit{machine learning}. As coined by \cite{mccarthy1989artificial}, AI is:
\change{
	\begin{definition}
		\label{bigdata_def}
		\textit{The science and engineering of making intelligent machines, especially computer systems by reproducing human intelligence through learning, reasoning and self-correction/adaption.} 
	\end{definition}
	AI uses \textit{intelligent agents} that perceive their environment and take actions that maximize their chance of successfully achieving their goals.}

\subsection{Machine Learning}
\noindent Machine learning (ML) is a subset of AI. ML aims to develop algorithms that can learn from historical data and improve the system with experience. In fact, by feeding the algorithms with data it is capable of changing its own internal programming to become better at a certain task. 
As coined by \cite{mitchell1990machine}:
\begin{definition}
	\label{bigdata_def}
	\textit{A computer program is said to learn from experience \textit{E} with respect to some
		class of tasks \textit{T} and performance measure \textit{P}, if its performance
		at tasks in \textit{T}, as measured by \textit{P}, improves
		with experience \textit{E}.} 
\end{definition}

ML experts focus on proving mathematical properties of new algorithms, compared to data mining experts who focus on understanding empirical properties of existing algorithms that they apply. \change{Within the broader picture of data science, ML is the step about taking the cleaned/transformed data and predicting future outcomes.}
Although ML is not a new field, with the significant increase of available data and the developments in computing and hardware technology ML has become one of the research hotspots in the recent years, in both academia and industry \cite{jiang2017machine}.

Compared to traditional signal processing approaches (e.g. estimation and detection), machine learning models are data-driven models; they do
not necessarily assume a data model on the underlying physical processes that generated the data. Instead, we may say they "\textit{let the data speak}", as they are able to infer or \textit{learn} the model. For instance, when it is complex to model the underlying physics that generated the wireless data, and given that there is sufficient amount of data available that may allow \tkcomm{to infer} the model \tkcomm{that} generalizes well beyond what is has seen, ML may outperform traditional signal processing and expert-based systems. \tkcomm{However, a representative amount and quality data is required.
The advantage of ML is that the resulting models are less prone to the modeling errors of the data generation process}.

\subsection{Deep Learning}
\noindent \change{Deep learning is a subset of ML, in which data is passed via multiple number of non-linear transformations to calculate an output.
	The term \textit{deep} refers to many steps in this case.
	A definition provided by \cite{lecun2015deep}, is:
	\begin{definition}
		\label{dl_def}
		\textit{Deep learning allows computational models that are composed of multiple processing layers to learn representations of data with multiple levels of abstraction.} 
	\end{definition}
}
A key advantage of deep learning over traditional ML approaches is that it can automatically extract high-level features from complex data. The learning process does not need to be designed by a human, which tremendously simplifies prior feature handcrafting \cite{lecun2015deep}. 

However, the performance of DNNs comes at the cost of the model's interpretability.
Namely, DNNs are typically seen as black boxes and there is lack of knowledge why they make certain decisions. Further, DNNs usually suffer \tkcomm{from complex hyper-parameters tuning, and finding their optimal configuration can be a challenge and time consuming.}
\tkcomm{Furthermore, training deep learning networks can be computationally demanding and requires advanced parallel computing
such as graphics processing units (GPUs).} Hence, when deploying deep learning models on embedded or mobile devices, considered should be the energy and computing constraints of the devices. 

\begin{figure}[t]
	\centering
	\includegraphics[width=0.5\textwidth]{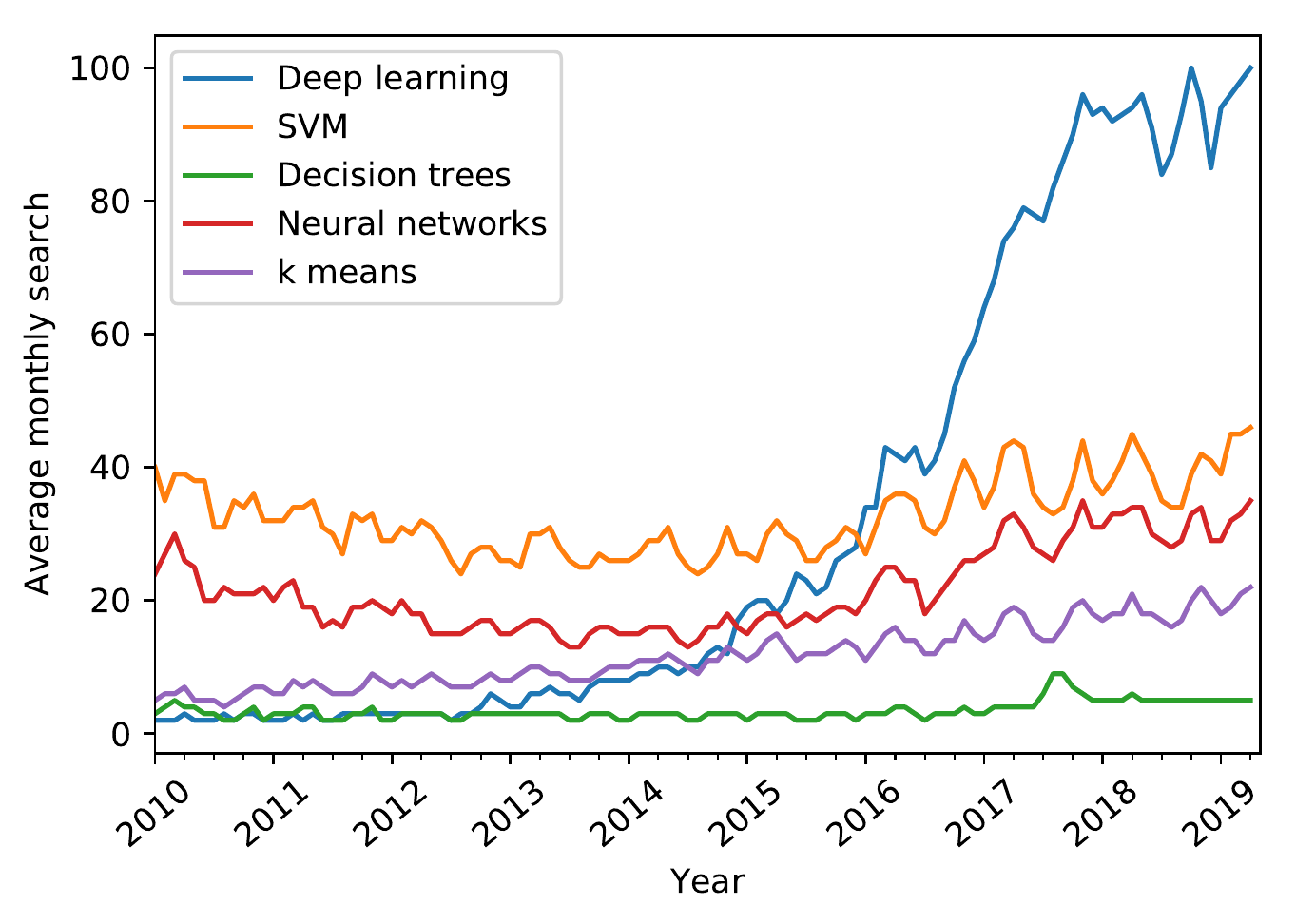}
	\caption{Google search trend showing increased attention in deep learning over the recent years}
	\label{fig:intro:DLtrend}
\end{figure}

There is a growing interest in \textit{deep learning} in the recent years. Figure \ref{fig:intro:DLtrend} demonstrates the growing interest in the field, showing the Google search trend from the past few years.

\section{Machine Learning Fundamentals}
\label{sec:4}
\noindent Due to their unpredictable nature, wireless networks are an interesting application area for data science because they are influenced by both, natural phenomena and man made artifacts. 
This section sets up the necessary fundamentals for the reader to understand the concepts of machine learning.

\subsection{\tkcomm{The Machine Learning Pipeline}}
\label{sec:ml_pipeline}
\tkcomm{Prior to applying machine learning algorithms to a wireless networking problem, the wireless networking problem needs to be first \textit{translated} into a data science problem.
In fact, the whole process from problem to solution may be seen as a machine learning pipeline consisting of several steps. 
\begin{figure*}[t]
	\centering
	\includegraphics[width=0.8\textwidth]{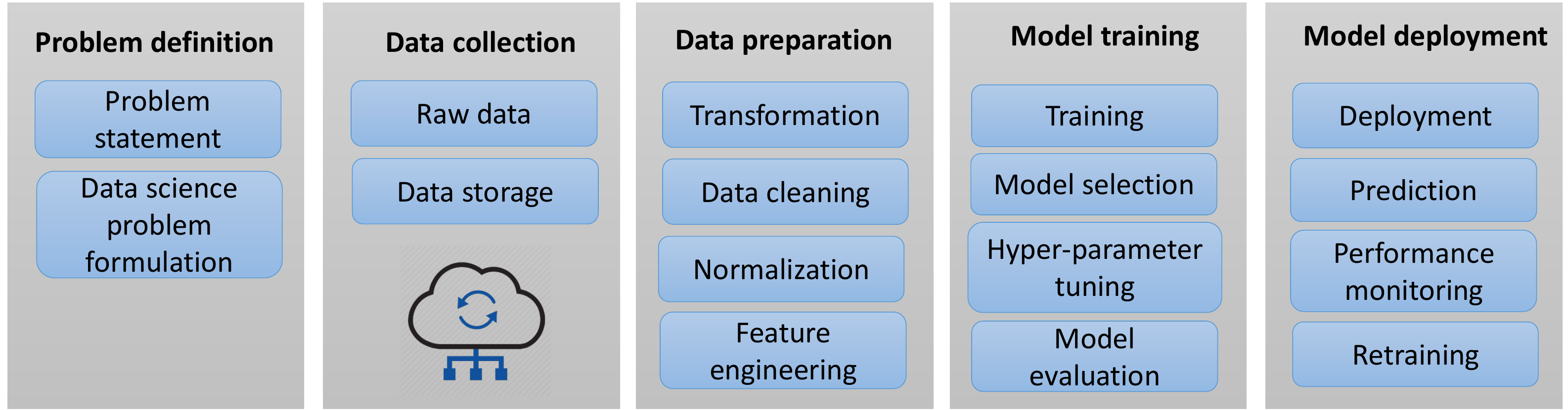}
	\caption{Steps in a machine learning pipeline}
	\label{fig:ML_pipeline}
\end{figure*}
Figure \ref{fig:ML_pipeline} illustrates those steps, which are briefly explained below:}
\begin{itemize}
	\item \tkcomm{\textbf{Problem definition.} In this step the problem is identified and translated into a data science problem. This is achieved by formulating the problem as a data mining task.
	Chapter \ref{subsec:learningProblems} further elaborates popular data mining methods such as classification and regression, and presents case studies of wireless networking problems of each type. In this way, we hope to help the reader understand how to formulate a wireless networking problem as a data science problem.
	\item \textbf{Data collection.} In this step, the needed amount of data to solve the formulated problem is identified and collected. The result of this step is \textit{raw data}.
	\item \textbf{Data preparation.} \tkcomm{After the problem is formulated and data is collected, the raw data is being preprocessed to be cleaned and transformed into a new space where each data pattern is represented by a vector, $\mathbf{x} \in \mathbb{R}^{n}$. This is known as the \textit{feature vector}, and its $n$ elements are known as \textit{features}. Through, the process of \textit{feature extraction} each pattern becomes a single point in a $n$-dimensional space, known as the \textit{feature space} or the \textit{input space}. Typically, one starts with some large value $P$ of features and eventually selects the $n$ most informative ones during the \textit{feature selection} process.} 
	\item \textbf{Model training.} After defining the feature space in which the data lays, one has to train a machine learning algorithm to obtain a model. This process starts by forming the \textit{training data} or  \textit{training set}. Assuming that $M$ feature vectors and corresponding known output values (sometimes called \textit{labels}) are available, the \textit{training set} $\mathcal{S}$ consists of $M$ input-output pairs ($(\mathbf{x}_i, y_i), i=1,...,M$) called \textit{training examples}, that is,
		\begin{equation} \label{dataset}
		\mathcal{S}=\{(\mathbf{x}_1, y_1),(\mathbf{x}_2, y_2), ..., (\mathbf{x}_M, y_M)\}\,,
		\end{equation}
		where $\mathbf{x}_i \in \mathbb{R}^{n}$, is the feature vector of  the $i$th observation,
		\begin{equation}
		\mathbf{x}_i=[x_{i1}, x_{i2}, ..., x_{in}]^{T}, i=1,...,M\,.
		\end{equation}
		The corresponding output values (\textit{labels}) to which $\mathbf{x}_i, i=1,...,M,$ belong, are 
		\begin{equation}
		\mathbf{y}=	[y_1, y_2, ..., y_M]^{T}\,.
		\end{equation}
	In fact, various ML algorithms are trained, tuned (by tuning their hyper-parameters) and the resulting models are evaluated based on standard performance metrics (e.g. mean squared error, precision, recall, accuracy, etc.) and the best performing model is chosen (i.e. model selection). 
	\item \textbf{Model deployment.} The selected ML model is deployed into a  practical wireless system where it is used to make predictions. For instance, given unknown raw data, first the feature vector $\mathbf{x}$ is formed, and then it is fed into the ML model for making predictions. Furthermore, the deployed model is continuously monitored to observe how it behaves in real world. To make sure it is accurate, it may be retrained.
}
	
\end{itemize}

Further below, the ML stage is elaborated in more detail.

\subsubsection{Learning the model}
\label{sec:learnmodel}
\noindent 
\tkcomm{Given a set $\mathcal{S}$, the goal of a machine learning algorithm is to \textit{learn} the mathematical model for $f$.
Thus, $f$ is some fixed but unknown function, that defines the relation between $\mathbf{x}$ and $y$, that is}
\begin{equation}
f: \mathbf{x} \rightarrow y \,.
\end{equation}

\tkcomm{The function $f$ is obtained by applying the selected learning method to the training set, $\mathcal{S}$, so that $f$ is a good estimator for new unseen data, i.e.,}
\begin{equation}
y \approx \hat{y}=\hat{f}(\mathbf{x}_{new})\,.
\end{equation}

In machine learning, $f$ is called the predictor, because its task is to \textit{predict} the outcome $y_i$ based on the input value of $\mathbf{x}_i$.
Two popular predictors are the \textit{regressor} and \textit{classifier}, described by:
\begin{equation}
f(x)=\left\{
\begin{array}{ll}
regressor \text{: } \text{ if } y \in \mathbb{R}	\\
classifier \text{: } \text{ if } y \in \{0,1\}	\\
\end{array}
\right.\,.
\end{equation}
In other words, when the output variable $y$ is \textit{continuous} or quantitative, the learning problem is a \textit{regression} problem. But, if $y$ predicts a discrete or \textit{categorical} value, it is a \textit{classification} problem.

\tkcomm{In case, when the predictor $f$ is parameterized by a vector $\boldsymbol{\uptheta} \in \mathbb{R}^{n} $, it describes a \textit{parametric} model.}
In this setup, the problem of estimating $f$ reduces down to one of estimating the parameters $\boldsymbol{\uptheta}=[\theta_1, \theta_2,...,\theta_n]^{T}$.
In most practical applications, the observed data are \tkcomm{noisy} versions of the expected values that would be obtained under ideal circumstances. These unavoidable \tkcomm{errors}, prevent the extraction of true parameters from the observations.
With this in regard, the generic data model may be expressed as
\begin{equation}
y=f(\mathbf{x})+\boldsymbol{\upepsilon}\,,
\end{equation}
where $f(\mathbf{x})$ is the model and $\boldsymbol{\upepsilon}$ are additive measurement errors and other discrepancies.
The goal of ML  is to find the input-output relation that will "best" match the noisy observations.
Hence, the vector $\boldsymbol{\uptheta}$ may be estimated by solving a \textit{(convex) optimization} problem. First, a \textit{loss} or \textit{cost function} $l(\mathbf{x}, \mathbf{y}, \boldsymbol{\uptheta})$ is set, which is a (point-wise) measure of the error 
between the observed data point $y_i$ and the model prediction $\hat{f}(\mathbf{x}_i)$ for each value of $\boldsymbol{\uptheta}$. However, $\boldsymbol{\uptheta}$ is estimated on the whole training set, $\mathcal{S}$, not just one example.
For this task, the average loss over all training examples called \textit{training loss}, $J$,  is calculated:
\begin{equation}
J(\boldsymbol{\uptheta}) \equiv J(S, \boldsymbol{\uptheta})=\frac{1}{m}\sum_{(\mathbf{x}_i,y_i) \in  S}l(\mathbf{x}_i,y_i, \boldsymbol{\uptheta})\,,
\end{equation}
where $\mathcal{S}$ indicates that the error is calculated on the instances from the training set and $i=1,...,m$.
The vector $\boldsymbol{\uptheta}$ that minimizes the training loss $J(\boldsymbol{\uptheta})$, that is
\begin{equation}
\operatornamewithlimits{argmin}\limits_{\boldsymbol{\uptheta} \in \mathbb{R}^n}{J(\boldsymbol{\uptheta})}\,,
\end{equation}
will give the desired model. Once the model is estimated, for any given input $\mathbf{x}$, the prediction for $y$ can be made with $\hat{y}=\boldsymbol{\uptheta}^T\mathbf{x}$.


\subsubsection{Learning the features}
\noindent The prediction accuracy of ML models heavily depends on the choice of the data representation or features used for training. For that reason, much effort in designing ML models goes into the composition of pre-processing and data transformation chains that result in a representation of the data that can support effective ML predictions. Informally, this is referred to as \textit{feature engineering}.
\textit{Feature engineering} is the process of extracting, combining and manipulating features by taking advantage of human ingenuity and prior expert knowledge to arrive at more representative ones. \tkcomm{The feature extractor $\phi$ transforms the data vector $\mathbf{d} \in \mathbb{R}^d$  into a new form, $\mathbf{x} \in \mathbb{R}^n$, $n <= d$, more suitable for making predictions, that is} 
\begin{equation} 
\phi(\mathbf{d}): \mathbf{d} \rightarrow \mathbf{x}\,.
\end{equation} 
For instance, the authors of \cite{liu2017wireless} engineered features from the RSSI (Received Signal Strength Indication) distribution to identify wireless signals.
The importance of feature engineering highlights the bottleneck of ML algorithms: their inability to automatically extract the discriminative information from data.
\textit{Feature learning} is a branch of machine learning that moves the concept of learning from "learning the model" to "learning the features". One popular feature learning method is \textit{deep learning}, in detail discussed in \ref{subsec:Deep Learning}. 

\subsection{\change{Types of learning paradigms}}
\begin{figure*}[t]
	\centering
	\includegraphics[width=0.75\textwidth]{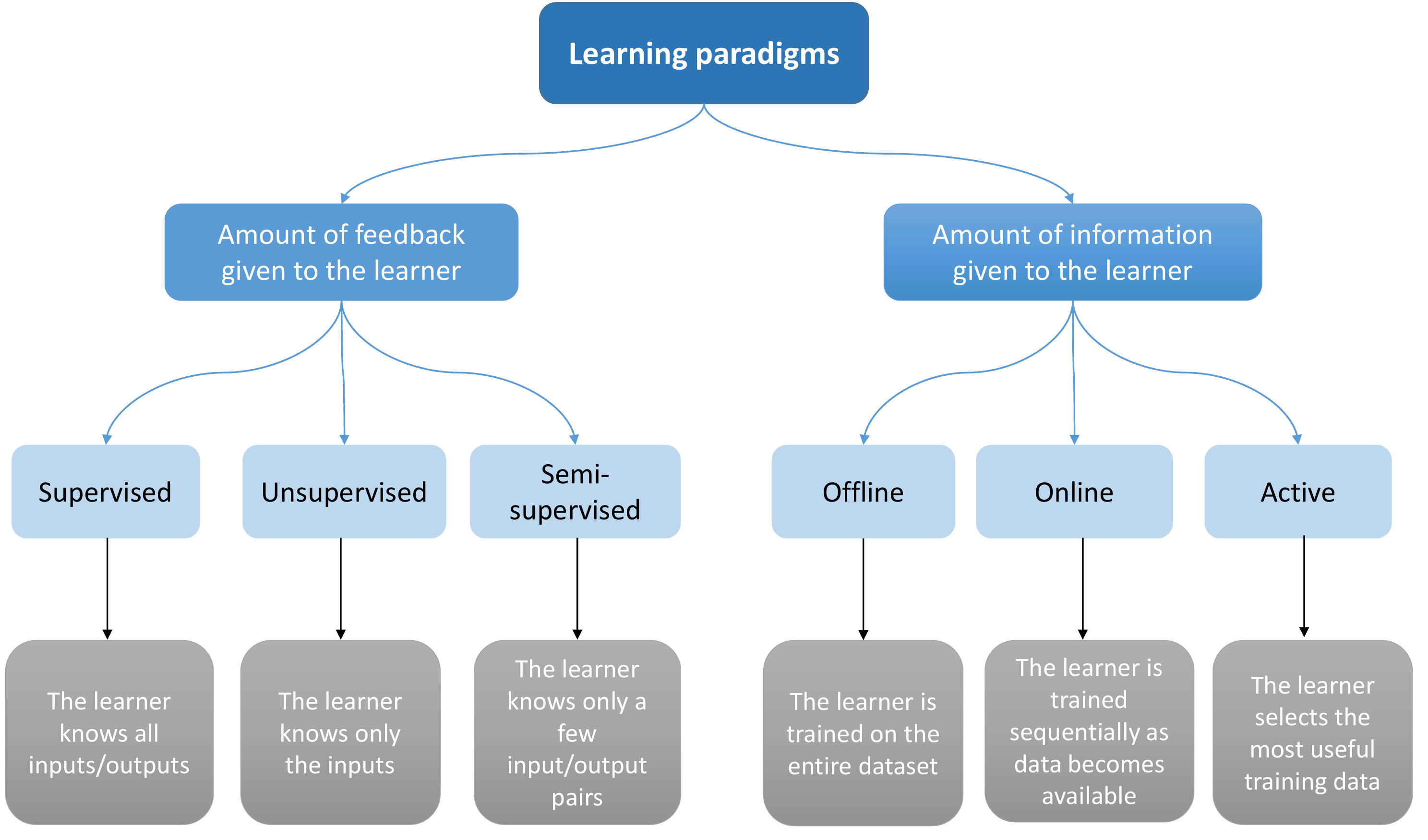}
	\caption{Summary of types of learning paradigms}
	\label{fig:learningParTable}
\end{figure*}

\noindent\change{This section discussed various types of learning paradigms in ML, summarized in Figure \ref{fig:learningParTable}}

\subsubsection{Supervised \textit{vs.} Unsupervised \textit{vs.} Semi-Supervised Learning}
\label{subsubsec:SupervisedVsUnsupervised}

\noindent Learning can be categorized by the amount of knowledge or feedback that is given to the learner as either supervised or unsupervised. 

\paragraph{Supervised Learning}
Supervised learning utilizes predefined inputs and known outputs to build a system model. The set of inputs and outputs forms the labeled training dataset that is used to teach a learning algorithm how to predict future outputs for new inputs that were not part of the training set. 
Supervised learning algorithms are suitable for wireless network problems where prior knowledge about the environment exists and data can be labeled.~For example, predict the location of a mobile node using an algorithm that is trained on signal propagation  characteristics (inputs) at known locations (outputs).~Various challenges in wireless networks have been addressed using supervised learning such as: medium access control \cite{sha2013self, kulkarni2009neural, kim2009bayesian, shen2008broadcast}, routing \cite{barbancho2006giving}, link quality estimation \cite{liu2014data, wang2007predicting}, node clustering in WSN \cite{ahmed2008cluster}, localization \cite{shareef2008localization, chagas2012approach, tran2008localization}, adding reasoning capabilities for cognitive radios \cite{tumuluru2010neural, baldo2008learning, tang2010artificial, hu2008signal, xu2006channel, petrova2010multi, huang2009design},  \textit{etc}.~Supervised learning has also been extensively applied to different types of wireless networks application such as: human activity recognition \cite{mannini2010machine, hong2006classification, lara2013survey, bulling2014tutorial, bao2004activity, bulling2012multimodal}, event detection \cite{yu2005real, bahrepour2009use, bahrepour2010distributed, zoha2014machine, khanafer2008automated}, electricity load monitoring \cite{ridi2014survey, chang2007load}, security \cite{branch2013network, kaplantzis2007detecting, kulkarni2009generalized}, \textit{etc}. Some of these works will be analyzed in more detail later.
\paragraph{Unsupervised Learning}
\tkcomm{Unsupervised learning} algorithms try to find hidden structures in unlabeled data. The learner is provided only with inputs without known outputs, while learning is performed by finding similarities in the input data. As such, these algorithms are suitable for wireless network problems where no prior knowledge about the outcomes exists, or annotating data (labelling) is difficult to realize in practice.
For instance, automatic grouping of wireless sensor nodes into clusters based on their current sensed data values and geographical proximity (without knowing a priori the group membership of each node) can be solved using unsupervised learning.
In the context of wireless networks, unsupervised learning algorithms are widely used for: data aggregation \cite{yoon2007clustered}, node clustering for WSNs \cite{he2009neural, kanungo2002efficient, liu2005dynamic, yoon2007clustered}, data clustering \cite{taherkordi2008communication, guo2009real, wang2005attribute}, event detection \cite{ma2013dynamic} and several cognitive radio applications \cite{clancy2011robust, shetty2009identifying}, dimensionality reduction \cite{o2017semi}, etc.
\paragraph{Semi-Supervised Learning}
Several mixes between the two learning methods exist and materialize into semi-supervised learning \cite{Witten:2005:DMP:1205860}. Semi-supervised learning is used in situations when a small amount of labeled data with a large amount of unlabeled data exists. It has great practical value because it may alleviate the cost of rendering a fully labeled training set, especially in situations where it is infeasible to label all instances. For instance, in human activity recognition systems where the activities change very fast so that some activities stay unlabeled or the user is not willing to cooperate in the data collection process, supervised learning might be the best candidate to train a recognition model \cite{guan2007activity, stikic2011weakly, huynh2006towards}.
Other potential use cases in wireless networks might be localization systems where it can alleviate the tedious and time-consuming process of collecting training data (calibration) in fingerprinting-based solutions \cite{pulkkinen2011semi} or semi-supervised traffic classification \cite{erman2007offline}, \textit{etc}.

\subsubsection{Offline \textit{vs.} Online \textit{vs.} Active Learning}

\noindent Learning can be categorized depending on the way the information is given to the learner as either offline or online learning. 
In offline learning the learner is trained on the entire training data at once, while in online learning the training data becomes available in a sequential order and is used to update the representation of the learner in each iteration.

\paragraph{Offline Learning}
Offline learning is used when the system that is being modeled does not change its properties dynamically. Offline learned models are easy to implement because the models do not have to keep on learning constantly, and they can be easily retrained  and redeployed in production.
For example, \mbox{in \cite{liu2011foresee}} a learning-based link quality estimator is implemented by deploying an offline trained model into the network stack of Tmote Sky wireless nodes. The model is trained based on measurements about the current status of the wireless channel that are obtained from extensive experiment setups from a wireless testbed. 

Another use cases are human activity recognition systems, where an offline trained  classifier  is deployed to
recognize actions from users. The classifier model can be trained  based on information extracted from raw measurements collected by sensors integrated in a smartphone, which is at the same time the central processing unit that implements the offline learned model for online activity recognition \cite{bin2012classification}. 

\paragraph{Online Learning}
Online learning is useful for problems where training examples arrive one at a time or when due to limited resources it is computationally infeasible to train over the entire dataset. For instance,  in \cite{bosman2015ensembles} a decentralized learning approach for anomaly detection in wireless sensor networks is proposed. The authors concentrate on detection methods that can be applied online (\textit{i.e}., without the need of an offline learning phase) and that are characterized by a limited computational footprint, so as to accommodate the stringent hardware limitations of WSN nodes.
Another example can be found in \cite{zhang2009adaptive}, where the authors propose an online outlier detection technique that can sequentially update the model and detect measurements that do not conform to the normal behavioral pattern of the sensed data, while maintaining the resource consumption of the network to a minimum.

\paragraph{Active Learning}
A special form of online learning is active learning where the learner first reasons about which examples would be most useful for training (taking as few examples as possible) and then collects those examples. Active learning has proven to be useful in situations when it is expensive to obtain samples from all variables of interest. For instance, the authors in \cite{dasarathy2016active} proposed a novel active learning approach (for graphical model selection problems), where the goal is to optimize the total number of scalar samples obtained by allowing the collection of samples from only subsets of the variables. This technique could for instance alleviate the need for synchronizing a large number of sensors to obtain samples from all the variables involved simultaneously.

Active learning has been a major topic in recent years in ML and an exhaustive literature survey is beyond the scope of this paper. We refer the reader for more details on active learning algorithms to  \cite{castro2008minimax,beygelzimer2010agnostic,hanneke2014theory}.

\subsection{Machine Learning Algorithms}
\label{ML_alg}
\noindent This section reviews popular ML algorithms used in wireless networks research.

\subsubsection{Linear Regression}
\noindent Linear regression is a supervised learning technique used
for modeling the relationship between a set of input (independent) variables ($\mathbf{x}$) and an output (dependent) variable ($y$), so that the output is a linear combination of the input variables: 
\begin{equation}
y=f(\mathbf{x}):=\theta_{0}+\theta_{1}x_{1}+...+\theta_{n}x_{n}+\upepsilon=\theta_{0}+\displaystyle\sum_{j=1}^{n} \theta_{j}x_{j}\,,
\end{equation}
\tkcomm{where $\mathbf{x}=[x_1,...x_n]^T$, and $\boldsymbol{\uptheta}=[\theta_{0},\theta_{1},...\theta_{n}]^T$ is the estimated parameter vector from a given training set $(y_i, \mathbf{x}_i)$, $i=1,2, ..., m$.}


\subsubsection{Nonlinear Regression}
Nonlinear regression is a supervised learning techniques which models the observed data by a function that is a nonlinear combination of the model parameters and one or more independent input variables. An example of nonlinear regression is the polynomial regression model defined by:
\begin{equation}
y=f(\mathbf{x}):=\theta_{0}+\theta_{1}x+\theta_{2}x^{2}+...+\theta_{n}x^{n}\,,
\end{equation}

\subsubsection{Logistic Regression}
Logistic regression \cite{freedman2009statistical} is a simple supervised learning algorithm widely used for implementing linear classification models, meaning that the models define smooth linear decision boundaries between different classes. At the core of the learning algorithm is the logistic function which is used to learn the model parameters and predict future instances.
The logistic function, $f(z)$, is given by $1$ over $1$
plus $e$ to the minus $z$, that is:
\begin{equation} 
f(z)=\frac{1}{1+e^{-z}}\,,
\end{equation}

\tkcomm{where, $z:=\theta_{0}+\theta_{1}x_1+ \theta_{2}x_2+...+\theta_{n}x_n$,} where  $x_1, x_2,...x_n$ are the independent (input) variables, that
we wish to use to describe or predict the dependent (output)
variable $y=f(z)$.

The range of $f(z)$ is between $0$ and $1$, regardless of the value of $z$, which makes it popular for classification tasks. Namely, the model is designed to describe a probability, which is always some number between 0 and 1.

\subsubsection{Decision Trees}
Decision trees (DT) \cite{maimon2008data} is a supervised learning algorithm that creates a tree-like graph or model that represents the possible outcomes or consequences of using certain input values. The tree consists of one root node, internal nodes called decision nodes which test its input against a learned expression, and leaf nodes which correspond to a final class or decision.
The learning tree can be used to derive simple decision rules that can be used for decision problems or for classifying future instances by starting at the root node and moving through the tree until a leaf node is reached where a class label is assigned.
However, decision trees can achieve high accuracy only if the data is linearly separable, \textit{i.e}., if there exists a linear hyperplane between the classes. Hence, constructing an optimal decision tree is NP-complete \cite{safavian1990survey}.

There are many algorithms that can form a learning tree such as the simple Iterative Dichotomiser~3 (ID3), its improved version  C4.5, \textit{etc}.

\subsubsection{Random Forest}
Random forests (RF) are \textit{bagged} decision trees. Bagging is a technique which involves training many classifiers and considering the average output of the ensemble. In this way, the variance of the overall ensemble classifier can be  greatly reduced. Bagging is often used with DTs as they are not very robust to errors due to variance in the input data.
Random forest are created by the following procedure:
\begin{algorithm}[tb]
	\caption{Random Forest}
	\KwIn{  Training set $D$}
	\KwOut{  Predicted value $h(x)$}
	Procedure:
	\begin{itemize}
		\item Sample $k$  datasets $D_1, ..., D_k$  from $D$ with replacement.
		\item For each $D_i$  train a decision tree classifier $h_i()$ to the maximum depth and when splitting the tree only consider a subset of features $l$. If $d$ is the number of features in each training example, the parameter $l<=d$ is typically set to $l=\sqrt{d}$.
		\item The ensemble classifier is then the mean or majority vote output decision out of all decision trees.
	\end{itemize}
	
\end{algorithm}

\begin{figure}[tb]
	\centering
	\includegraphics[width=0.5\textwidth]{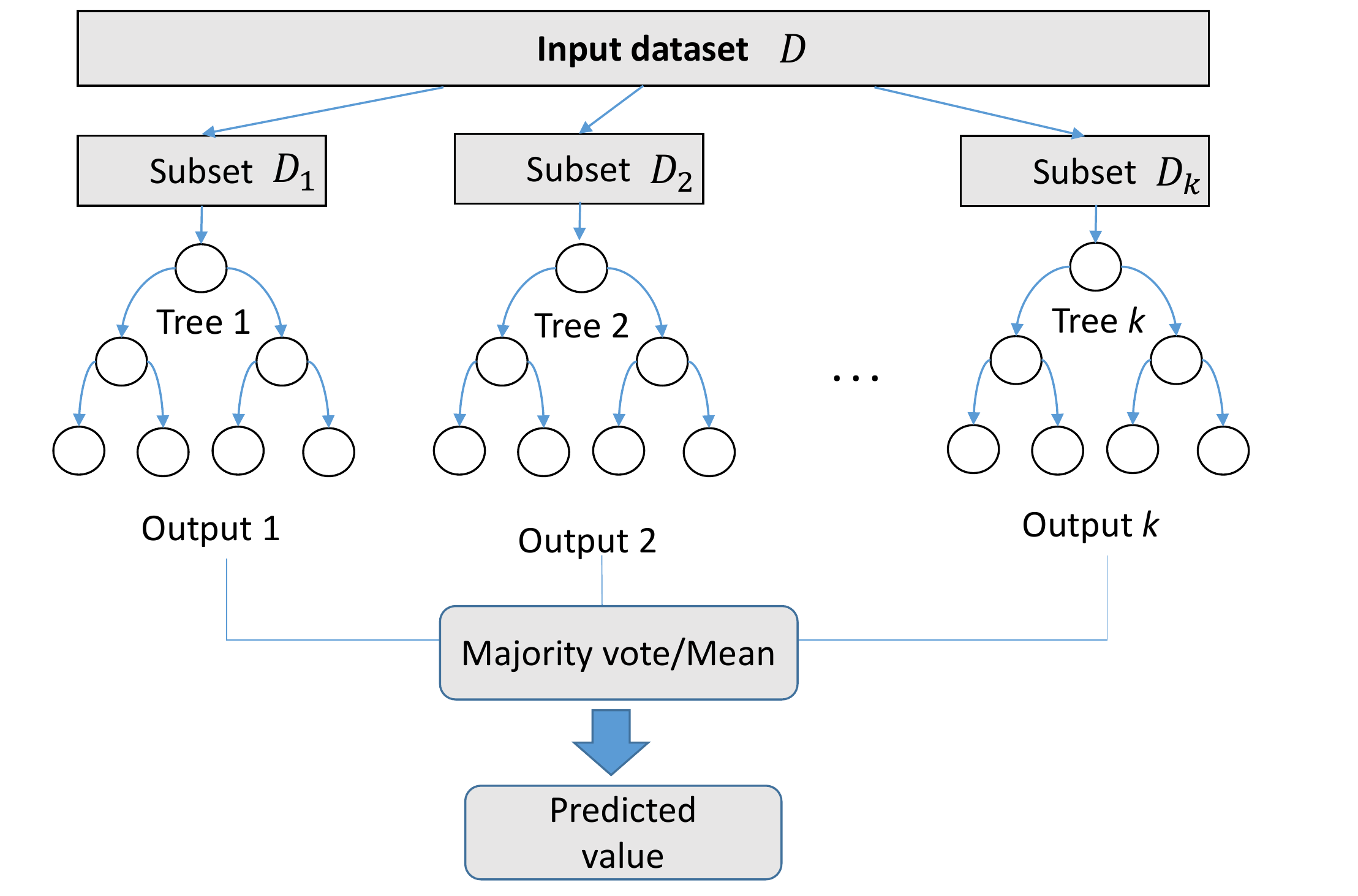}
	\caption{Graphical formulation for Random Forest}
	\label{fig:ch2:RF_alg}
\end{figure}

Figure \ref{fig:ch2:RF_alg} illustrates this process.

\subsubsection{SVM}
Support Vector Machine (SVM) \cite{vapnik1998statistical} is a learning algorithm that solves classification problems by first mapping the input data into a higher-dimensional feature space in which it becomes linearly separable by a hyperplane, which is used for classification. In Support vector regression, this hyperplane is used to predict the continuous value output.
The mapping from the input space to the high-dimensional feature space is non-linear, which is achieved using \textit{kernel} functions. Different kernel functions comply best for different application domains. The most common kernel functions used in SVM are: linear kernel, polynomial kernel and basis kernel function (RBF), given as:
\begin{equation}
\resizebox{0.45\textwidth}{!}{$
	k(x_i,x_j)=x_{i}^{T}x_{j} \\
	k(x_i,x_j)=(x_{i}^{T}x_{j}+1)^{d} \\
	k(x_i,x_j)=e^{-\frac{(x_i - x_j)^{2}}{\sigma^{2}}} \\
	$}\,,
\end{equation}
where $\sigma$ is a user defined parameter.

\subsubsection{k-NN}
\textit{k} nearest neighbors (k-NN) \cite{larose2005k} is a learning algorithm that can solve classification and regression problems by looking into the distance (closeness) between input instances. It is called a non-parametric learning algorithm because, unlike other supervised learning algorithms, it does not learn an explicit model function from the training data. Instead, the algorithm simply memorizes all previous instances and then predicts the output by first searching the training set for the \textit{k} closest instances and then: (i) for classification-predicts the majority class amongst those \textit{k} nearest neighbors, while (ii) for regression-predicts the output value as the average of the values of its \textit{k} nearest neighbors.
Because of this approach, k-NN is considered a form of instance-based or \mbox{memory-based learning.}

k-NN is widely used since it is one of the simplest forms of learning. It is also considered as \textit{lazy} learning as the learner is passive until a prediction has to be performed, hence no computation is required until performing the prediction task. 
The pseudocode for k-NN \cite{dunham2006data} is summarized in Algorithm \ref{k-nn_pseudocode}.

\begin{algorithm}
	\caption{\tkcomm{k-NN}}
	\KwIn{$(y_i, \mathbf{x}_i)$: Training set, $i=1,2, ..., m$; $s$: unknown sample}
	\KwOut{  Predicted value $f(\mathbf{x})$}
	\textbf{Procedure:}
	\For{$i \gets 1$ to $m$}   {
		$Compute$ $distance$ $d(\mathbf{x}_i, s)$}
	\begin{enumerate}
		\item $Compute$ $set$ $I$ $containing$ $indices$ $for$ $the$ $k$ $smallest$ $distances$ $d(\mathbf{x}_i, s)$
		\item $f(\mathbf{x}) \gets$ majority label/mean value for $\{y_i \text{ where } i \in I\}$
	\end{enumerate}
	\Return {$f(\mathbf{x})$}
	\label{k-nn_pseudocode}
\end{algorithm}

\subsubsection{k-Means}
\begin{figure}[bt]
	\centering
	\includegraphics[width=0.5\textwidth]{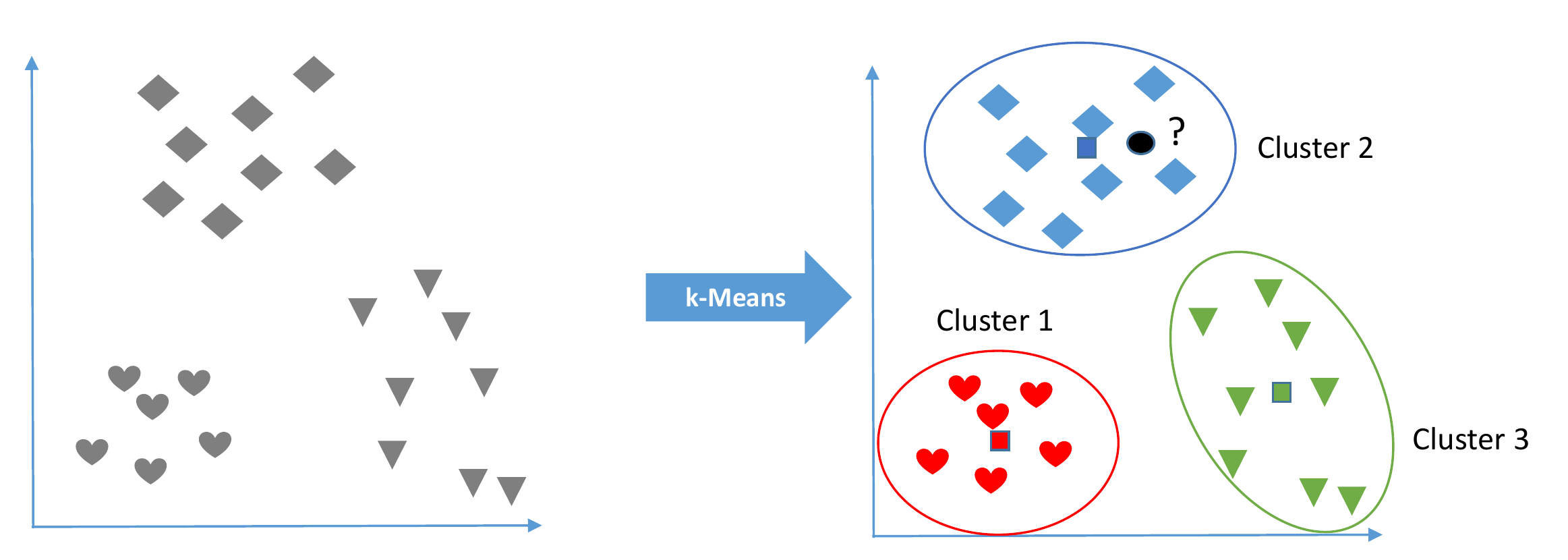}
	\caption{Graphical formulation for k-Means}
	\label{fig:ch2:kmeans_alg}
\end{figure}

\textit{k}-Means is an unsupervised  learning algorithm \tkcomm{used for clustering problems. The goal is to assign a number of points, $x_1, .., x_m$ into \textit{K} groups or clusters}, so that the resulting intra-cluster similarity is high, while the inter-cluster similarity low. The similarity is measured with respect to the mean value of the data points in a cluster.
Figure \ref{fig:ch2:kmeans_alg} illustrates an example of k-means clustering, where $K=3$ and the input dataset consisting of two features with data points plotted along the $x$ and $y$ axis.

On the left side of Figure \ref{fig:ch2:kmeans_alg} are data points before k-means is applied, while on the right side are the identified 3 clusters and their centroids represented with squares.

The pseudocode for k-means \cite{dunham2006data} is summarized in Algorithm \ref{k-means_pseudocode}.

\begin{algorithm}
	\caption{\tkcomm{k-means}}	
	\KwIn{$K$: The  number of desired clusters; $X=\{x_1, x_2,...,x_m\}$: Input dataset with $m$ data points}
	\KwOut{A set of $K$ clusters}	\textbf{Procedure:}
	\begin{enumerate}
		\item Set the cluster centroids $\mu_k$, $k=1, ...,K$ to arbitrary values;
		\item \While{no change in $\mu_k$} {
			\begin{enumerate}	
				\item (Re)assign each item $x_i$ to the 
				cluster with the closest centroid.
				\item Update $\mu_k$, $k=1, ...,K$, as 
				the mean value of the data points in each cluster. 
			\end{enumerate}	
		}
	\end{enumerate}	
	\Return {$K$ clusters}
	\label{k-means_pseudocode}
\end{algorithm}

\subsubsection{Neural Networks}
\begin{figure}[b]
	\centering
	\includegraphics[width=0.5\textwidth]{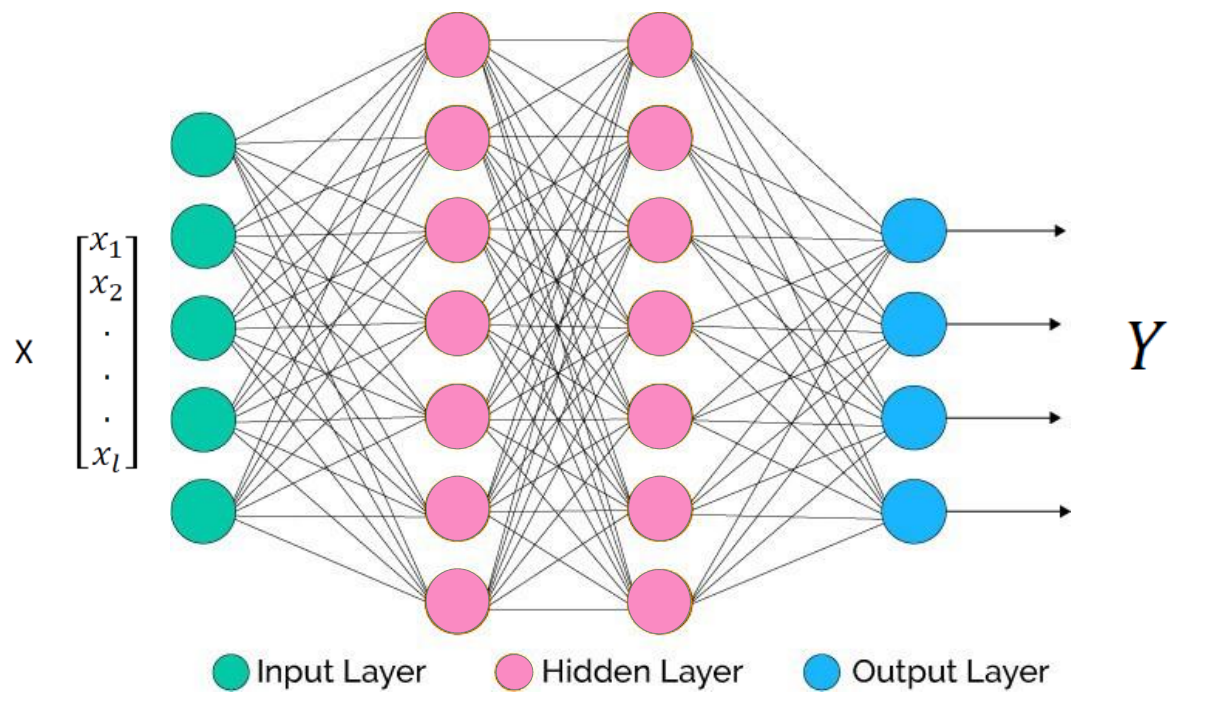}
	\caption{Graphical formulation for Neural networks}
	\label{fig:ch2:nn_alg}
\end{figure}

Neural Networks (NN) \cite{haykin2009neural} or artificial neural networks (ANN) is a supervised learning algorithm inspired on the working of the brain, that is typically used to derive complex, non-linear decision boundaries for building a classification model, but are also suitable for training regression models when the goal is to predict real-valued outputs (regression problems are explained in \mbox{Section \ref{subsubsec:regression}).} Neural networks are known for their ability to identify complex trends and detect complex non-linear relationships among the input variables at the cost of higher computational burden.
A neural network model consists of one input, a number of hidden layers and one output layer, as shown on Figure \ref{fig:ch2:nn_alg}.

The formulation for a single layer is as follow:
\begin{equation} \label{eq:spatconv}
y=g(\mathbf{w}^T\mathbf{x}+b)\,,
\end{equation}
where $\mathbf{x}$ is a training example input, and $y$ is the layer output, $\mathbf{w}$ are the layer weights, while $b$ is the bias term.

The input layer corresponds to the input data variables. Each hidden layer consists of a number of processing elements called neurons that process its inputs (the data from the previous layer) using an activation or transfer function that translates the input signals to an output signal, $g()$. Commonly used activation functions are: unit step function, linear function, sigmoid function and the hyperbolic tangent function. The elements between each layer are highly connected by connections that have numeric weights that are learned by the algorithm.
The output layer outputs the prediction (\textit{i.e}., the class) for the given inputs and according to the interconnection weights defined through the hidden layer. The algorithm is again gaining popularity in recent years because of new techniques and more powerful hardware that enable training complex models for solving complex tasks. 
In general, neural networks  are said to be able to approximate any function of interest when tuned well, which is why they are considered as universal approximators \cite{Hornik1989}.

\paragraph{Deep neural networks}
\label{subsec:Deep Learning}
\change{Deep neural networks are a special type of NNs consisting of multiple layers able to perform feature transformation and extraction. Opposed to a traditional NN, they have the potential to alleviate  manually extracting features, which is a process that depends much on prior knowledge and domain expertise \cite{jia2016deep}.} 

\change{Various deep learning techniques exist, including: deep neural networks (DNN), convolutional neural networks (CNN), recurrent neural networks (RNN) and deep belief networks (DBN), which have shown success in various fields of science including computer vision, automatic speech recognition, natural language processing, bioinformatics, \textit{etc}, and  increasingly also in wireless networks.}

\begin{figure*}[bt]
	\centering
	\includegraphics[width=0.85\textwidth]{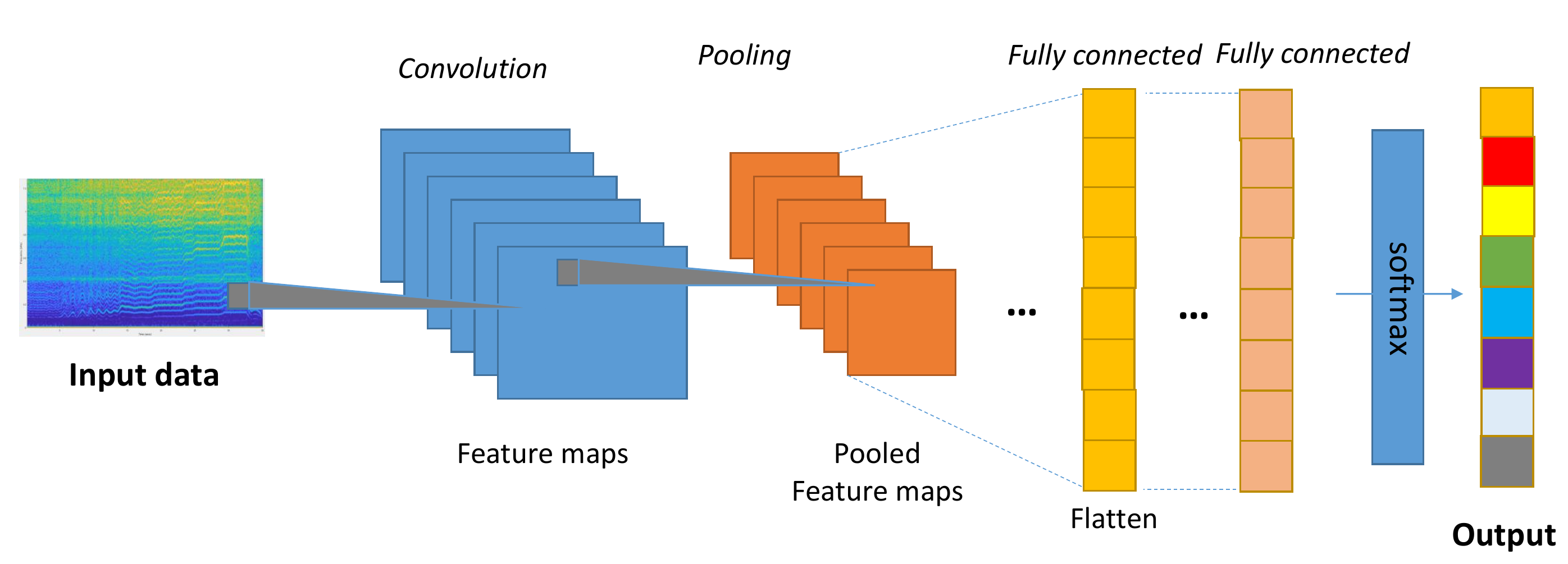}
	\caption{Graphical formulation of Convolutional Neural Networks}
	\label{fig:ch2:cnn}
\end{figure*}

\paragraph{Convolutional neural networks}
Convolutional neural networks (CNN) perform feature learning via non-linear transformations implemented as a series of nested \textit{layers}. 
The input data is a multidimensional data array, called \textit{tensor}, that is presented at the \textit{visible layer}.
This is typically a grid-like topological structure, e.g. time-series data, which can be seen as a 1D grid taking samples at regular time intervals, pixels in images with a 2D layout, a 3D structure of videos, etc. 
Then a series of \textit{hidden layers} extract several abstract features. Hidden layers consist of a series of convolution, pooling and
fully-connected layers, as shown on Figure \ref{fig:ch2:cnn}.

Those layers are "hidden" because their values are not given. Instead, the deep learning model must determine which data representations are useful for explaining the relationships in the observed data.
Each convolution layer consists of several \textit{kernels} (i.e. filters) that perform a \textit{convolution} over the input; therefore, they are also referred to as \textit{convolutional layers}.
Kernels are feature detectors, that convolve over the input and produce a transformed version of the data at the output. 
Those are banks of finite impulse response \textit{filters} as seen in signal processing, just learned on a hierarchy of layers.
The filters are usually  multidimensional arrays of parameters that are learnt by the learning algorithm \cite{Goodfellow-et-al-2016} through a  training process called \textit{backpropagation}.

For instance, given a two-dimensional input $x$, a two-dimensional kernel $h$ computes the 2D convolution by
\begin{equation} \label{eq:spatconv}
\resizebox{0.45\textwidth}{!}{$(x*h)_{i,j}=x[i,j]*h[i,j]=\sum_{n}\sum_{m} x[n,m] \cdot h[i-n][j-m]$}\,,
\end{equation}
i.e. the dot product between their weights and a small region they are connected to in the input. 

After the convolution, a bias term is added and a point-wise nonlinearity $g$ is applied, forming a \textit{feature map} at the filter output. If we denote the $l$-th feature map at a given convolutional layer as $\mathbf{h}^l$, whose filters are determined by the coefficients or \textit{weights} $\mathbf{W}^l$, the input $\mathbf{x}$ and the bias $b_l$, then the feature map $h^l$ is obtained as follows
\begin{equation} \label{eq:featuremap}
{h^l}_{i,j}=g({(W^l*x)}_{ij}+{b}_l)\,,
\end{equation}
where $*$ is the 2D convolution defined by Equation \ref{eq:spatconv}, while $g(\cdot)$ is the \textit{activation function}. 


Common activation functions encountered in deep neural networks are the \textit{rectifier}  that is defined as
\begin{equation} \label{eq:relu}
g(x)=x^{+}=max(0,x)\,,
\end{equation}
the hyperbolic tangent function, \textit{tanh},  $g(x)=tanh(x)$, that is defined as
\begin{equation} \label{eq:tanh}
tanh(x)=\frac{2}{1+e^{-2x}}-1\,,
\end{equation}
and the \textit{sigmoid} activation, $g(x)=\sigma(x)$, defined as
\begin{equation} 
\sigma(x)=\frac{1}{1+e^{-x}}\,.
\end{equation}

The \textit{sigmoid} activation is rarely used because its activations saturate at either tail of $0$ or $1$ and they are not centered at $0$ as is the \textit{tanh}. The \textit{tanh} normalizes the input to the range $[-1, 1]$, but compared to the rectifier its activations saturate which causes unstable gradients.
Therefore, the \textit{rectifier} activation function is typically used for CNNs. Kernels using the rectifier are called \textit{ReLU} (Rectified Linear Unit) and have shown to greatly accelerate the convergence during the training process compared to other activation functions. 
They also do not cause vanishing or exploding of gradients in the optimization phase when minimizing the cost function. In addition, the ReLU simply thresholds the input, $x$, at zero, while other activation functions involve expensive operations.

In order to form a richer representation of the input signal, commonly, multiple filters are stacked so that each hidden layer consists of multiple \textit{feature maps}, $\{h^{(l)}, l = 0,...,L\}$ (e.g., $L=64,128,...$, etc).
The number of filters per layer is a tunable parameter or \textit{hyper-parameter}. Other tunable parameters are the filter size, the number of layers, etc. The selection of values for hyper-parameters may be quite difficult, and finding it commonly is much an art as it is science. An optimal choice may only be feasible by trial and error.
The filter sizes are selected according to the input data size so as to have the right level of  “granularity” that can create abstractions at the proper scale.
For instance, for a 2D square matrix input, such as spectrograms, common choices are $3 \times 3$, $5\times 5$, $9 \times 9$, etc. For a \textit{wide} matrix, such as a real-valued representation of the complex I and Q samples of the wireless signal in $\mathbb{R}^{2 \times N}$, suitable filter sizes may be $1 \times 3$, $2 \times 3$, $2 \times 5$, etc.

After a convolutional layer, a \textit{pooling} layer may be used to merge semantically similar features into one. In this way, the spatial size of the representation is reduced which reduces the amount of parameters and computation in the network. Examples of pooling units are \textit{max pooling} (computes the maximum value of a local patch of units in one feature map), \textit{neighbouring pooling} (takes the input from patches that are shifted
by more than one row or column, thereby reducing the dimension of the representation and creating an invariance to small shifts and distortions, etc.

The penultimate layer in a CNN consists of \textit{neurons} that are fully-connected with all feature maps in the preceding layer. Therefore, these layers are called \textit{fully-connected} or \textit{dense} layers.
The very last layer is a \textit{softmax} classifier, which computes the \textit{posterior} probability of each class label over $K$ classes as 
\begin{equation} \label{eq:softmax}
\hat{y_i}=\frac{e^{z_i}}{\sum_{j=1}^{K}e^{z_j}}, \text{  } i=1,...,K
\end{equation}
That is, the scores $z_i$ computed at the output layer, also called \textit{logits}, are translated into probabilities.
A loss function, $l$, is calculated on the last fully-connected layer
that measures the difference between the estimated probabilities, $\hat{y_i}$, and the one-hot encoding of the true class labels, $y_i$.
The CNN parameters, $\boldsymbol{\uptheta}$, are obtained by minimizing the loss function on the training set $\{ x_i, y_i\}_{i \in S}$ of size $m$,
\begin{equation}
\operatornamewithlimits{min}\limits_{\boldsymbol{\uptheta}} \sum_{i \in  S}l(\hat{y_i}, y_i)\,,
\end{equation}
where $l(.)$ is typically the  mean squared error $l(y,\hat{y})=\| y-\hat{y} \|_2^2$ or the \textit{categorical cross-entropy} 
$l(y,\hat{y})={\sum_{i=1}^{m} y_ilog(\hat{y_i})}$ for which a minus sign is often added in front to get the negative \textit{log-likelihood}.
Then the softmax classifier is trained by solving an optimization problem that minimizes the loss function.
The optimal solution are the network parameters that fully describe the CNN model. That is $\hat{\boldsymbol{\uptheta}}=  \operatornamewithlimits{argmin}\limits_{\boldsymbol{\uptheta}}J(S,\boldsymbol{\uptheta})$.

Currently, there is no consensus about the choice of the optimization algorithm. The most successful optimization algorithms seem to be: stochastic gradient descent (SGD), RMSProp, Adam, AdaDelta, etc. For a comparison on these, we refer the reader to \cite{schaul2013unit}. 

To control over-fitting, typically regularization is used in combination with \textit{dropout}, which is a new extremely effective technique that "drops out" a random set of activations in a layer. Each unit is retained with a fixed probability $p$,
typically chosen using a validation set, or set to $0.5$ which
has shown to be close to optimal for a wide range of applications \cite{srivastava2014dropout}.

\begin{figure}[bt]
	\centering
	\includegraphics[width=0.5\textwidth]{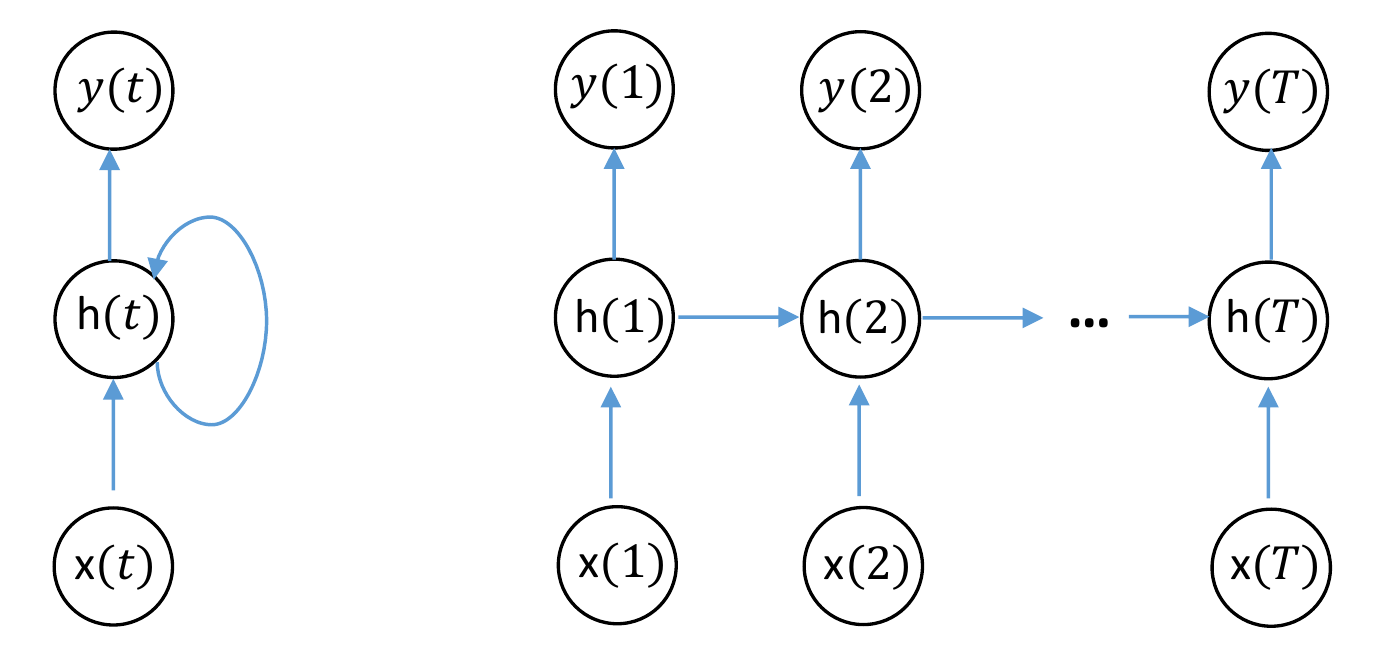}
	\caption{Graphical formulation of Recurrent Neural Networks}
	\label{fig:ch2:rnn_alg}
\end{figure}

\paragraph{Recurrent neural networks}
Recurrent neural networks (RNN) \cite{rumelhart1988learning} are a type of neural networks where connections between nodes form a directed graph along a temporal sequence. They are called recurrent because of the recurrent connections between the hidden units. This is mathematically denoted as:
\begin{equation} \label{eq:featuremap}
\mathbf{h}^{(t)}=f(\mathbf{h}^{(t-1)},\mathbf{x}^{(t)};\theta)
\end{equation} 
where function $f$ is the activation output of a single unit, $\mathbf{h}^(i)$ are the state of the hidden units at time $i$, $\mathbf{x}^(i)$ is the input from the sequence at time index $i$, $\mathbf{y}^(i)$ is the output at time $i$,
while $\theta$ are the network weight parameters used to compute the activation at all indices.
Figure \ref{fig:ch2:rnn_alg} shows a graphical representation of RNNs.

The left part of Figure  \ref{fig:ch2:rnn_alg} presents the "folded" network, while the right part the "unfolded" network with its recurrent connections propagating information forward in time. An activation functional is applied in the hidden units and the $softmax$ may be used to calculate the prediction.

There are various extensions of RNNs. A popular extension are LSTMs, which augment the traditional RNN model by adding a self loop on the state of the network to better “remember” relevant information over longer periods in time.

\section{\change{Data Science} Problems in Wireless Networks}
\label{subsec:learningProblems}
\noindent \change{The ultimate goal of data science is to extract knowledge from data, \textit{i.e}., turn data into real value \cite{fayyad1996kdd}. At the heart of this process are severe algorithms that can \textit{learn} from and make predictions on data, i.e. machine learning algorithms.
In the context of wireless networks, learning is a mechanism that enables context awareness and intelligence capabilities in different aspects of wireless communication.  Over the last years, it has gained popularity due to its success in enhancing network-wide performance (\textit{i.e}. QoS) \cite{yau2012reinforcement}, facilitating intelligent behavior by adapting to complex and dynamically changing (wireless) environments \cite{CI2009} and its ability to add automation for realizing concepts of self-healing and self-optimization \cite{khatib2016self}. During the past years, different data-driven approaches have been studied in the context of: mobile ad hoc networks \cite{forster2007machine}, wireless sensor networks \cite{kulkarni2011computational}, wireless body area networks \cite{lara2013survey}, cognitive radio networks \cite{thilina2013machine, clancy2007applications} and cellular networks \cite{anagnostopoulos2009predicting}. These approaches are focused on addressing various topics including: medium access control \cite{sha2013self, esteves2015cooperative}, routing \cite{liu2011foresee, Liu2014temporal}, data aggregation and clustering \cite{yoon2007clustered, chen2013tw}, localization \cite{vanheel2011automated, tennina2014wsn4qol}, energy harvesting communication \cite{blasco2013learning}, spectrum sensing \cite{hu2008signal, huang2009design}, \textit{etc}.}

\tkcomm{As explained in section \ref{sec:ml_pipeline},} prior to applying ML to a wireless networking problem, the problem needs to be first \change{formulated as} an adequate data mining method.

\change{This section explains the following methods:
	\begin{itemize}
		\item Regression
		\item Classification
		\item Clustering
		\item Anomaly Detection
	\end{itemize}
}
\tkcomm{For each problem type, several wireless networking case studies are discussed together with the ML algorithms that are applied to solve the problem.}

\subsection{Regression}
\label{subsubsec:regression}
\begin{figure}[b]
	\centering
	\includegraphics[width=0.35\textwidth]{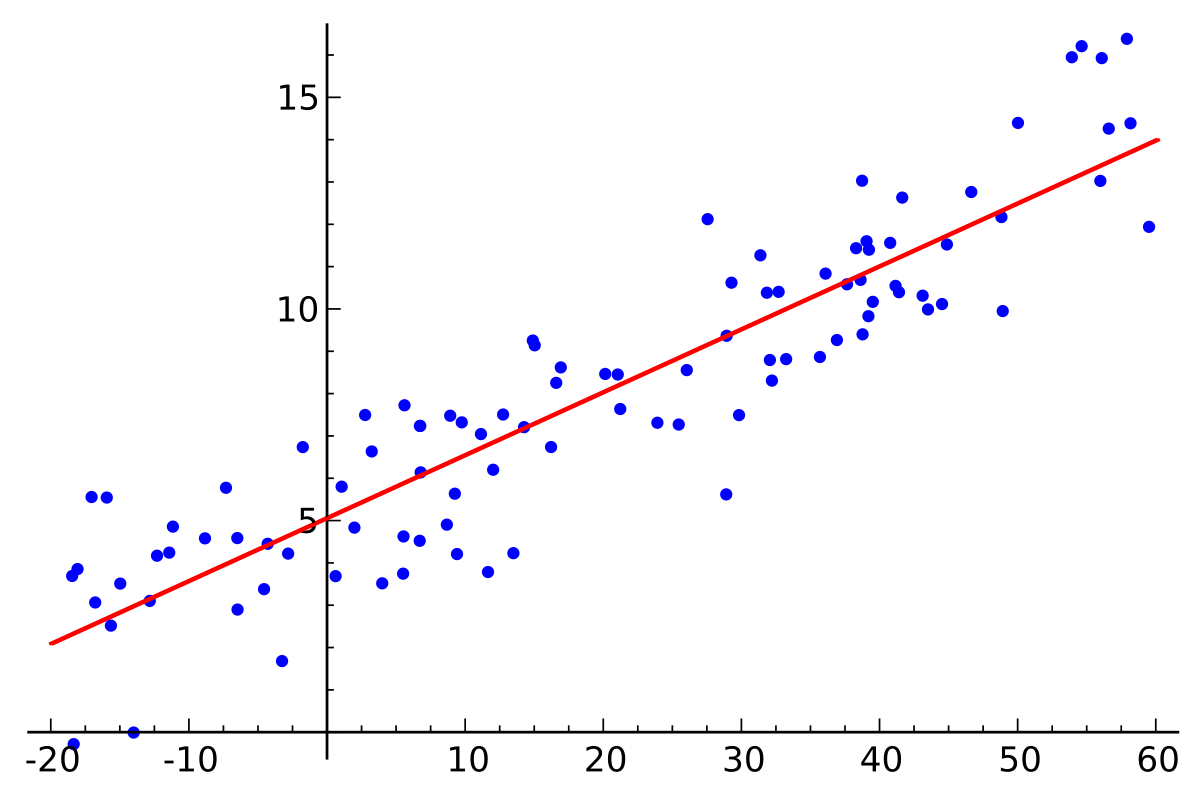}
	\caption{Illustration of regression.}
	\label{fig:regression}
\end{figure}

\noindent Regression is a data mining method that is suitable for problems that aim to predict a real-valued output variable, $y$, as illustrated on Figure \ref{fig:regression}.
\tkcomm{Given a training set, $\mathcal{S}$, the goal is to estimate a function, $f$, whose graph fits the data. Once the function $f$ is found, when an unknown point arrives, it is able to predict the output value. This function $f$ is known as the \textit{regressor}.}

Depending on the function representation, regression techniques are typically categorized into linear and non-linear \mbox{regression algorithms}, \tkcomm{as explained in section \ref{ML_alg}. For example, linear channel equalization in wireless communication can be seen as a regression problem.}

\subsubsection{Regression Example 1: Indoor localization}

In the context of wireless networks, linear regression is frequently used to derive an empirical log-distance model for the radio propagation characteristics as a linear mathematical relationship between the RSSI, usually in dBm, and the distance. This model can be used in RSSI-based indoor localization algorithms to estimate the distance towards each fixed node (\textit{i.e}., anchor node) in the ranging phase of the algorithm \cite{vanheel2011automated}.

\subsubsection{Regression Example 2: Link Quality estimation}
Non-linear regression techniques are extensively used for modeling the relation between the PRR (Packet Reception Rate) and the RSSI, as well as between PRR and the Link Quality Indicator (LQI), to build a mechanism to estimate the link quality based on observations (RSSI, LQI) \cite{levis2006rssi}.

\subsubsection{Regression Example 3: Mobile traffic demand prediction}
The authors in \cite{bega2019deepcog} use ML to optimize network resource allocation in mobile networks. Namely, each base station observes the traffic of a particular network slice in a mobile network. Then, a CNN model uses this information to predict the capacity required to accommodate the future traffic demands for services associated to each network slice. In this way, each slice gets optimal resources allocated.

\subsection{Classification}
\label{subsubsec:classification}

\begin{figure}[tb]
	\centering
	\includegraphics[width=0.3\textwidth]{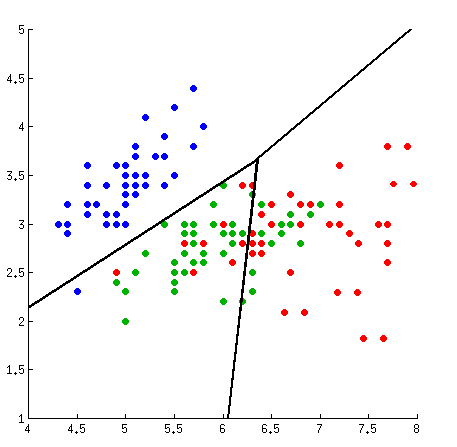}
	\caption{Illustration of classification.}
	\label{fig:classification}
\end{figure}

\noindent A classification problem tries to understand and predict discrete values or categories. The term classification comes from the fact that it predicts the class membership of a particular input instance, as shown on Figure \ref{fig:classification}. \tkcomm{Hence, the goal in classification is to assign an unknown \textit{pattern} to one out of a number of classes that are considered to be known. For example, in digital communications, the process of demodulation can be viewed as a classification problem. Upon receiving the modulated transmitted signal, which has been impaired by propagation effects (i.e.the channel) and noise, the receiver has to decide which data symbol (out of a finite set) was originally transmitted. 
}

Classification problems can be solved by supervised learning approaches, that aim to model boundaries between sets (\textit{i.e}., classes) of similar behaving instances, based on known and labeled (\textit{i.e}., with defined class membership) input values. 
There are many learning algorithms that can be used to classify data including decision trees, k-nearest neighbours, logistic regression, support vector machines, neural networks, convolutional neural networks, etc.

\subsubsection{Classification Example 1: Cognitive MAC layer}
We consider the problem of designing an adaptive MAC layer as an application example of decision trees in wireless networks. In \cite{sha2013self} a self-adapting MAC layer is proposed. It is composed of two parts: (i) a reconfigurable MAC architecture that can switch between different MAC protocols at run time, and (ii) a trained MAC engine that selects the most suitable MAC protocol for the current network condition and application requirements. The MAC engine is solved as a classification problem using a decision tree classifier which is learned based on: (i) two types of input variables which are (1) network conditions reflected through the RSSI statistics (\textit{i.e}., mean and variance), and (2) the current traffic pattern monitored through the Inter-Packet Interval (IPI) statistics (\textit{i.e}., mean and variance) and application requirements (\textit{i.e}., reliability, energy consumption and latency), and (ii) the output which is the MAC protocol that is to be predicted and selected.

\subsubsection{Classification Example 2: Intelligent routing in WSN}
Liu \textit{et al}. \cite{liu2011foresee} improved multi-hop wireless routing by creating a data-driven learning-based radio link quality estimator.~They investigated whether machine learning algorithms (e.g., logistic regression, neural networks) can perform better than traditional, manually-constructed, pre-defined estimators such as STLE (Short-Term Link Estimator) \cite{alizai2009bursty} and 4Bit (Four-Bit) \cite{fonseca2007four}.~Finally, they selected logistic regression as the most promising model for solving the following classification problem: predict whether the next packet will be successfully received, \textit{i.e}., output class is 1, or lost, \textit{i.e}., output class is 0, based on the current wireless channel conditions reflected by statistics of the PRR, RSSI, SNR and LQI.

While in \cite{liu2011foresee} the authors used offline learning to do prediction, in their follow-up work \cite{Liu2014temporal}, they went a step further and both training and prediction were performed online by the nodes themselves using logistic regression with online learning (more specifically the stochastic gradient descent online learning algorithm). The advantage of this approach is that the learning and thus the model, adapt to changes in the wireless channel, that could otherwise be captured only by re-training the model offline and updating the implementation on the node.

\subsubsection{Classification Example 3: Wireless Signal Classification}
ML has been extensively used in cognitive radio applications to perform signal classification. For this purpose, typically flexible and reconfigurable SDR (software defined radio) platforms are used to sense the environment to obtain information about the wireless channel conditions and users' requirements, while intelligent algorithms build the cognitive learning engine that can make decisions on those reconfigurable parameters on SDR \mbox{(e.g., carrier} frequency, transmission power, modulation scheme). 

In \cite{hu2008signal, huang2009design, ramon2009signal} SVMs are used as the machine learning algorithm to classify signals among a given set of possible modulation schemes. For instance, Huang \textit{et al}. \cite{huang2009design} identified four spectral correlation features that can be extracted from signals for distinction of different modulation types. Their trained SVM classifier was able to distinguish six modulation types with high accuracy: AM, ASK, FSK, PSK, MSK and QPSK.


\subsection{Clustering}
\label{subsubsec:clustering}
\begin{figure}[t]
	\centering
	\includegraphics[width=0.3\textwidth]{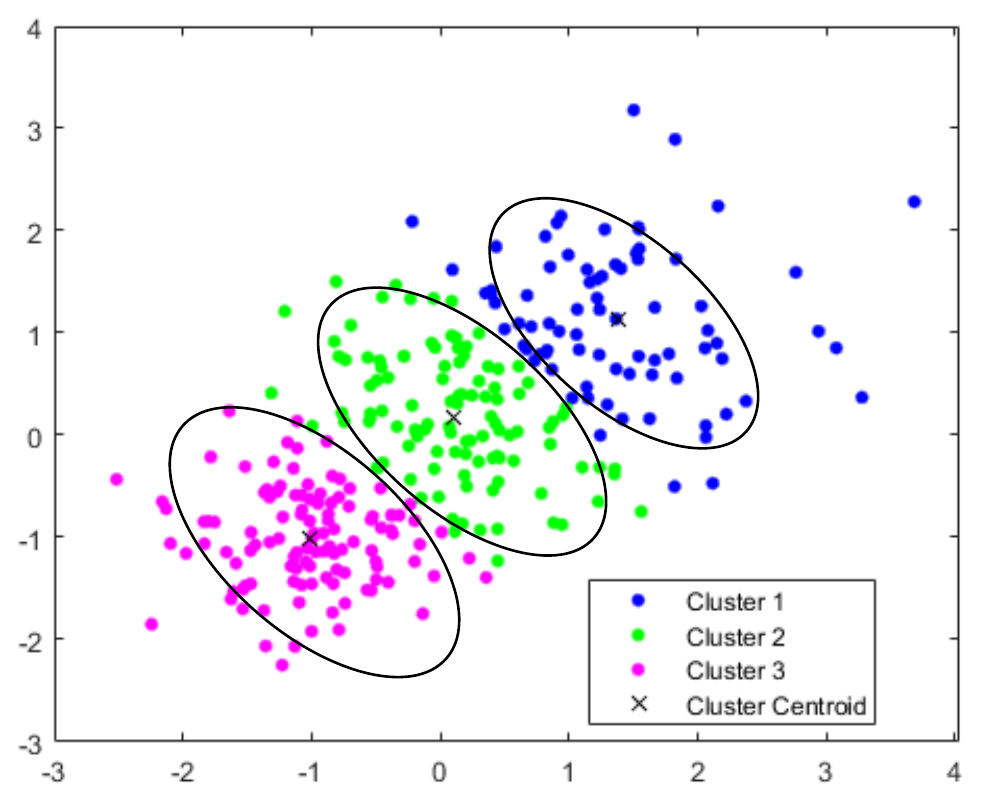}
	\caption{Illustration of clustering.}
	\label{fig:clustering}
\end{figure}

\noindent Clustering is a data mining method that can be used for problems where the goal is to group sets of similar instances into clusters, as shown on Figure \ref{fig:clustering}.

Opposed to classification, it uses \textit{unsupervised} learning, which means that the input dataset instances used for training are not labeled, \textit{i.e}., it is unknown to which group they belong. The clusters are determined by inspecting the data structure and grouping  objects that are similar according to some metric.
Clustering algorithms are widely adopted in wireless sensor networks, where they have found use for grouping sensor nodes into clusters to satisfy scalability and energy efficiency objectives, and finally elect the head of each cluster. A significant number of node clustering algorithms tends to be proposed for WSNs \cite{abbasi2007survey}. However, these \textit{node} clustering algorithms typically do not use the data science clustering techniques directly. Instead, they exploit \textit{data} clustering techniques to find data correlations or similarities between data of neighboring nodes, that can be used to partition sensor nodes into clusters.

Clustering can be used to solve other types of problems in wireless networks like anomaly detection, \textit{i.e}., outliers detection, such as intrusion detection or event detection, for different data pre-processing tasks, cognitive radio application (e.g., identifying wireless systems \cite{shetty2009identifying}), \textit{etc}.~There are many learning algorithms that can be used for clustering, but the most commonly used is \textit{k}-Means.
Other popular clustering algorithms include hierarchical clustering methods such as single-linkage, complete-linkage, centroid-linkage; graph theory-based clustering such as highly connected subgraphs (HCS),  cluster affinity search technique (CAST); kernel-based clustering as is support vector clustering (SVC), \textit{etc}. A novel two-level clustering algorithm, namely TW-\textit{k}-means, has been introduced by Chen \textit{et al}. \cite{chen2013tw}. For a more exhaustive list of clustering algorithms and their explanation we refer the reader to \cite{xu2005survey}.
Several clustering approaches have shown promise for designing efficient data aggregation for more efficient communication strategies in low power wireless sensor networks constrained. Given the fact that the most of the energy on the sensor nodes is consumed while the radio is turned on, \textit{i.e}., while sending and receiving data \cite{kimura2005survey}, clustering may help to aggregate data in order to reduce transmissions and hence energy consumption.

\subsubsection{Clustering Example 1: Summarizing sensor data}
In \cite{taherkordi2008communication} a distributed version of the \textit{k}-Means clustering algorithm was proposed for clustering data sensed by sensor nodes.~The clustered data is summarized and sent towards a sink node.~Summarizing the data ensures to reduce the communication transmission, processing time and power consumption of the sensor nodes.

\subsubsection{Clustering Example 2: Data aggregation in WSN}
In \cite{yoon2007clustered} a data aggregation scheme is proposed for in-network data summarization to save energy and reduce computation in wireless sensor nodes. The proposed algorithm uses clustering to form clusters of nodes sensing similar values within a given threshold. Then, only one sensor reading per cluster is transmitted which lowered extremely the number of transmissions in the wireless sensor network.

\subsubsection{Clustering Example 3: Radio signal identification}
The authors of \cite{o2017semi} use clustering to separate
and identify radio signal classes without to alleviate the need of using explicit class labels on examples of radio signals.
First, dimensionality reduction is performed on signal examples to transform the signals into a space suitable for signal clustering. Namely, given an
appropriate dimensionality reduction, signals are turned into a space where
signals of the same or similar type have a low distance
separating them while signals of differing types are separated
by larger distances. 
Classification of radio signal types in such a space then becomes
a problem of identifying clusters and associating a label with each
cluster. The authors used the DBSCAN clustering algorithm \cite{ester1996density}.

\subsection{Anomaly Detection}
\label{subsubsec:anomalyDet} 
\noindent Anomaly detection (changes and deviation detection) is used when the goal is to identify unusual, unexpected or abnormal system behavior. This type of problem can be solved by supervised or unsupervised learning depending on the amount of knowledge present in the data (\textit{i.e}., whether it is labeled or unlabeled, respectively). 

\begin{figure}[tb]
	\centering
	\includegraphics[width=0.3\textwidth]{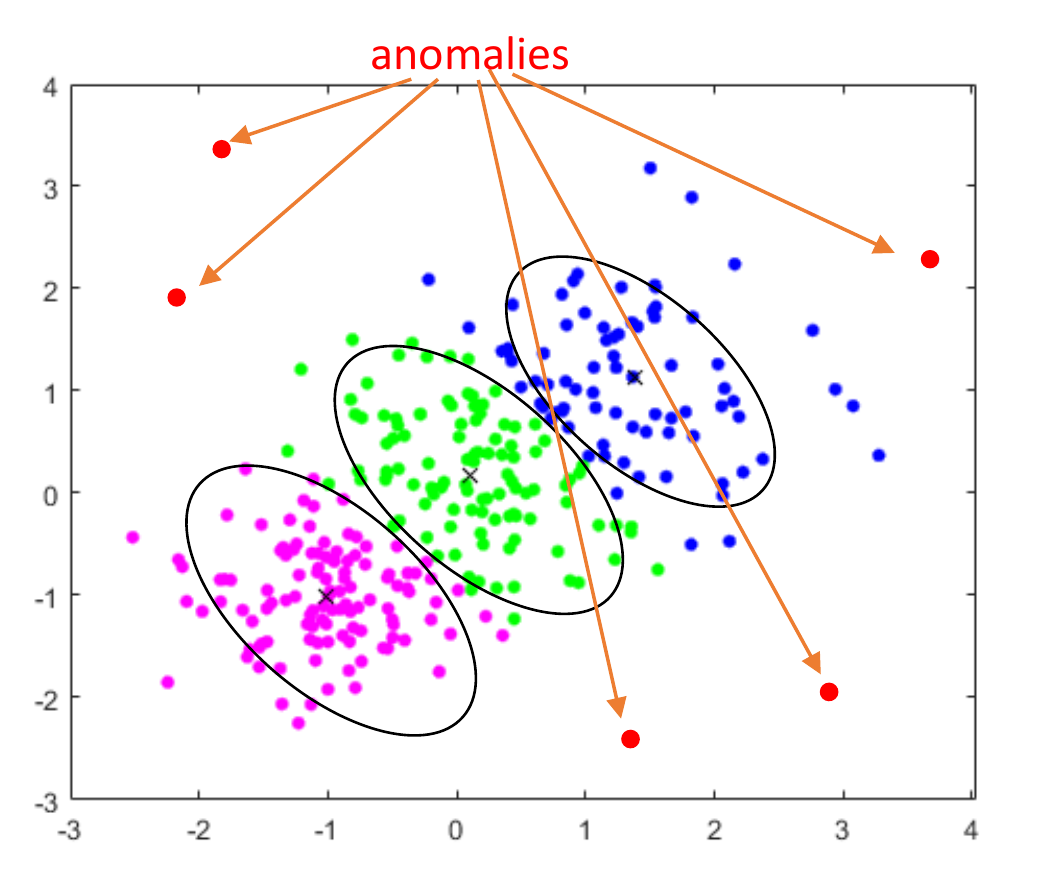}
	\caption{Illustration of anomaly detection.}
	\label{fig:anomaly}
\end{figure}

Accordingly, classification and clustering algorithms can be used to solve anomaly detection problems. Figure \ref{fig:anomaly} illustrates anomaly detection.
A wireless example is the detection of suddenly occurring phenomena, such as the identification of suddenly disconnected networks due to interference or incorrect transmission power settings.~It is also widely used for outliers detection in the pre-processing phase \cite{zhang2010outlier}. Other use-case examples include intrusion detection, fraud detection, event detection in sensor networks, \textit{etc}.

\subsubsection{Anomaly Detection Example 1: WSN attack detection}
WSNs have been target of many types of DoS attacks. The goal of DoS attacks in WSNs is to transmit as many packets as possible whenever the medium is detected to be idle. This prevents a legitimate sensor node from transmitting their own packets. To combat a DoS attack, a secure MAC protocol based on neural networks has been proposed \mbox{in \cite{kulkarni2009neural}}. The NN model is trained to detect an attack by monitoring variations of following parameters: collision rate $R_c$, average waiting time of a packet in MAC buffer $T_w$, arrival rate of RTS packets $R_{RTS}$. An anomaly, \textit{i.e}., attack, is identified when the monitored traffic variations exceeds a preset threshold, after which the WSN node is switched off temporarily. The results is that flooding  the network with untrustworthy data is prevented by blocking only affected sensor nodes.

\subsubsection{Anomaly Detection Example 2: System failure and intrusion detection}
In \cite{bosman2015ensembles} online learning techniques have been used to incrementally train a neural network for in-node anomaly detection in wireless sensor network. More specifically, the Extreme Learning Machine algorithm \cite{huang2004extreme} has been used to implement classifiers that are trained online on resource-constrained sensor nodes for detecting anomalies such as: system failures, intrusion, or unanticipated behavior of the environment.

\subsubsection{Anomaly Detection Example 3: Detecting wireless spectrum anomalies}
In \cite{rajendran2019unsupervised} wireless spectrum anomaly detection has been studied. The authors use Power Spectral Density (PSD) data to detect and localize anomalies (e.g. unwanted signals in the licensed band or the absence of an expected signal) in the wireless spectrum using a combination of Adversarial autoencoders (AAEs), CNN and LSTM.

\section{Machine Learning for Performance Improvements in Wireless Networks}
\label{sub:mlsurvey}

Obviously, machine learning is increasingly used in wireless networks \cite{jiang2017machine}.
After carefully looking at the literature, \change{we identified} two distinct categories or objectives where machine learning empowers wireless networks with the ability to learn and infer from data and extract patterns:

\begin{itemize}
	\item \textbf{Performance improvements} of the wireless networks \change{based on performance indicators and environmental insights (e.g. about the radio medium) as input, acquired from the devices. 
		These approaches exploit ML to generate patterns or make predictions, which are used to modify operating parameters at the PHY, MAC and network layer. }
	\item \textbf{Information processing} of data generated by wireless devices \change{at the application layer. 
		This category covers various applications such as: IoT environmental monitoring applications, activity recognition, localization, precision agriculture, etc.}
\end{itemize}

\change{This section presents tasks related to each of the aforementioned objectives achieved via ML and discusses existing work in the domain.}
First, the works are broadly summarized in tabular form in Table \ref{tab:intro:twoMLobjective}, followed by a detailed discussion of the most important works in each domain.

\change{The focus of this paper is on the first category related to ML for performance improvement of wireless networks, therefore, a comprehensive overview of the existing work addressing problems pertaining to communication performance by making use of ML techniques is presented in the forthcoming subsection. These works provide a promising direction towards solving problems caused by the proliferation of wireless devices, networks and technologies in the near future, including: problems with interference (co-channel interference, inter-cell interference, cross technology interference, multi user interference, etc.), non-adaptive modulation scheme, static non-application cognizant MAC, etc.}

\begin{table*}[hbt!]
	\centering
	\large
	\begin{adjustbox}{width=0.8\textwidth}
		\centering
		{\renewcommand{\arraystretch}{2}%
			\begin{tabular}{p{4.8cm} p{5cm} l p{4cm}}
				\toprule
				\textbf{Goal} & \textbf{Scope/Area}  & \textbf{Example of problem} & \textbf{References} \bigstrut[b]\\
				\hline
				\hline
				\toprule
				
				\multirow{6}{*}{\textbf{Performance Improvement}} 
				
				& \multirow{2}{*}{Radio spectrum analysis}  & $\bullet$ AMR & \cite{wong2004automatic,wang2009recognitionAMR,tabatabaei2010svm,hassan2010automatic,aubry2011cumulants,popoola2011novel,aslam2012automatic,valipour2012automatic,popoola2013effect,satija2015automatic,o2017learning,hassanpour2016automatic,o2016convolutional,kim2016deep,peng2017modulation,ali2017automatic,liu2017deep,o2017introduction,o2017semi,west2017deep,karra2017modulation,hauser2017signal,paisana2017context,zhao2018specific,youssef2018machine,jagannath2018artificial,		o2018over,san2018evaluating,mossaddeep2018,rajendran2018deep,tang2018digital,zhang2018automaticuav,jagannath2018artificial,peng2018modulation,duan2018automatic,meng2018automatic,wu2018deep,wu2018vhf,zhang2018automatic,li2018generative,li2018radioCNN,yashashwi2019learnable,sadeghi2019adversarial,ramjee2019fast}\\
				
				& & $\bullet$ Wireless interference identification & \cite{o2017spectral,bitar2017wireless,schmidt2017wireless,han2017spectrum,grunau2018multi,sun2018learning,yi2018interference,maglogiannis2019enhancing,rajendran2019unsupervised,soto2018detection,lee2018resource,awe2018spatio,han2017spectrum,lee2019deepsensing,sadeghi2019adversarial,fontaine2019towards}\\ 
				
				\cline{2-4}
				
				& \multirow{3}{*}{MAC analysis} & $\bullet$ MAC identification & \cite{yang2010mac,hu2012mac,hu2014mac,rajab2015energy}\\
				
				& & $\bullet$ Wireless interference identification & \cite{rayanchu2011airshark,hermans2013sonic,zheng2014zisense,hithnawi2015tiim} \\
				
				& & $\bullet$ Spectrum prediction & \cite{mennes2018neural,wang2018deep,xu2018realtime,yu2019deep,zhang2019neural,mennes2019deep} \\ \cline{2-4}
				
				& \multirow{2}{*}{Network prediction}  & $\bullet$ Network performance prediction & \cite{baldo2009neural}, \cite{liu2011foresee}, \cite{liu2012talent}, \cite{Liu2014temporal}, \cite{adeel2015critical}, \cite{pierucci2016neural}, \cite{sha2013self}, \cite{qiao2016mac}, \cite{kulin2017poster}, \cite{akbas2018neural}, \cite{al2017new}\\
				
				& & $\bullet$ Network traffic prediction & \cite{baldo2009neural,liu2011foresee,liu2012talent,Liu2014temporal,adeel2015critical,pierucci2016neural,sha2013self,qiao2016mac,kulin2017poster,akbas2018neural,al2017new}\\
				\midrule
				\multirow{8}{*}{\textbf{Information processing}}  
				& \multirow{4}{*}{IoT Infrastructure monitoring} & $\bullet$ Smart farming & \\ 
				
				& & $\bullet$ Smart mobility & \cite{lottes2017uav,sa2018weednet,strohbach2015towards,nguyen2016traffic} \\
				& & $\bullet$ Smart city & \cite{rathore2015efficient,amato2017deep,mittal2016spotgarbage,gillis2016nonintrusive} \\
				& & $\bullet$ Smart grid & \\ \cline{2-4}
				
				& Wireless security   &  Device fingerprinting & \cite{lv2017device,jafari2018iot,merchant2018deep,thing2017ieee,uluagac2013passive,bezawada2018behavioral,riyaz2018deep} \\ \cline{2-4}
				
				& \multirow{2}{*}{Wireless localization} & $\bullet$ Indoor & 				
				\multirow{2}{*}{\cite{wang2015phasefi,wang2016csi,wang2017csi,wang2015deepfi,wang2017device,zhang2016deeplearning}} \\
				& & $\bullet$ Outdoor & \\ \cline{2-4}
				& Activity recognition & Via wireless signals  & \cite{zeng2016wiwho,zhao2018through,lv2019robust,shahzad2018augmenting,wang2015understanding}    \\
				\bottomrule	
		\end{tabular}}%
	\end{adjustbox}
	\caption{An overview of the applications of machine learning in wireless networks}
	\label{tab:intro:twoMLobjective}
\end{table*}

\subsection{\change{Machine Learning Research} for Performance improvement}

Data generated during monitoring of wireless networking infrastructure (e.g. throughput, end-to-end delay, jitter, packet loss, etc.)
and by the wireless sensor devices (e.g. spectrum monitoring) and analyzed by ML techniques
has the potential to optimize wireless networks configurations, thereby improving end-users’ QoE.
Various works have applied ML techniques for gaining insights that can help improve the network performance.
Depending on the type of data used as input for ML algorithms, we first categorize the researched literature into three types, summarized in Table \ref{tab:intro:twoMLobjective}: 

\begin{itemize}
	\item \textbf{Radio spectrum analysis}
	\item \textbf{Medium access control (MAC) analysis}
	\item \textbf{Network prediction}
\end{itemize}

\begin{figure*}[th]
	\centering
	\includegraphics[width=0.8\textwidth]{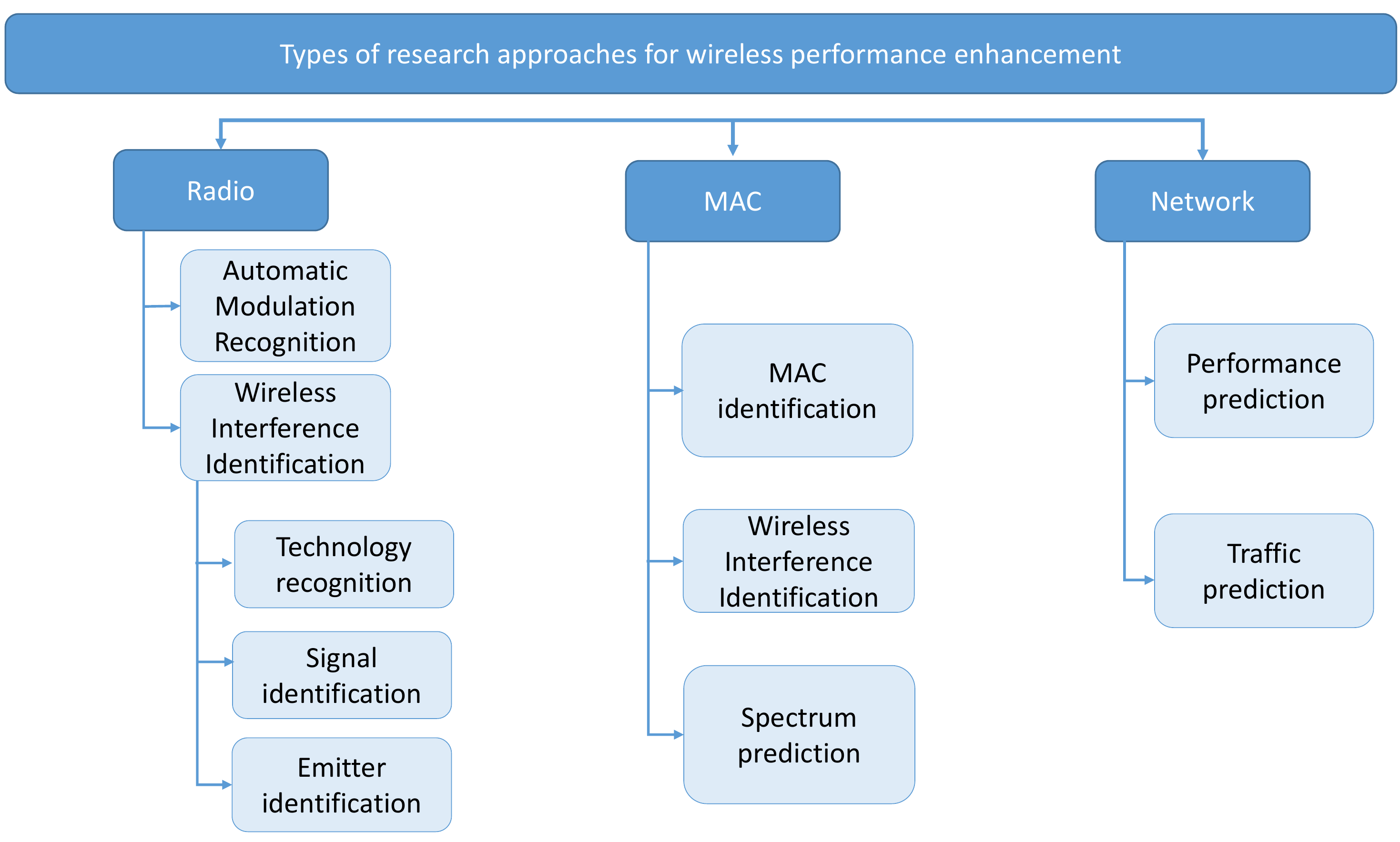}
	\caption{Types of research approaches for performance improvement of wireless networks}
	\label{fig:wpeApproaches}
\end{figure*}

Furthermore, within each of the above categories, we identified several classes of research approaches illustrated in Figure 
\ref{fig:wpeApproaches}. In what follows, the work in these directions is reviewed.

\subsubsection{Radio spectrum analysis}
Radio spectrum analysis refers to investigating wireless data sensed by the wireless devices to infer the radio spectrum usage. Typically, the goal is to detect unused spectrum portions in order to share it with other coexisting users within the network without exorbitant interference with each other. 
Namely, as wireless devices become more pervasive throughout society the available radio spectrum, which is a scarce resource, will contain more non-cooperative signals than seen before. Therefore, collecting information about the signals within the spectrum of interest is becoming ever more important and complex. This has motivated the use of ML for analyzing the signals occupying the radio spectrum.

Perhaps the most prevalent task related to radio spectrum analysis solved using ML is \textbf{\textit{automatic modulation recognition}} (AMR). Other related radio spectrum analysis tasks which employ ML techniques include \textit{technology recognition} (TR) and \textit{signal identification} (SI) methods. Typically, the goal is to detect the presence of signals that may cause interference so as to decide on a interference mitigation strategy. Therefore, we introduce those approached as \textbf{\textit{wireless interference identification}} (WII) tasks.

\textbf{\textit{Automatic modulation recognition.}} AMR plays a key role in various civilian and military applications, where \textit{friendly} signals shall be securely transmitted and received, whereas hostile signals must be located, identified and jammed. 
In short, the goal of this task is to recognize the type of modulation scheme an emitter is using to modulate its transmitting signal based on raw samples of the detected signal at the receiver side. This information can provide insight about the type of communication systems and emitters present in the radio environment.

Traditional AMR algorithms were classified into likelihood-based (LB) approaches \cite{ozdemir2013hybrid}, \cite{ozdemir2015asynchronous}, \cite{wimalajeewa2015distributed} and feature-based (FB) approaches \cite{azzouz2013automatic}, \cite{alharbi2012automatic}.  
LB approaches are based on detection theory (i.e. hypotesis testing) \cite{chepuri2011performance}. They can offer good performance and are considered optimal classifiers, however they suffer high computation complexity. Therefore, FB approaches were developed as suboptimal classifiers suitable for practical use. Conventional FB approaches heavily rely on expert knowledge, which may perform well for specialized solutions, however they are poor in generality and are time-consuming. Namely, in the preprocessing phase of designing the AMR algorithm, traditional FB approaches extracted complex hand engineered features (e.g. some signal parameters) computed from the raw signal and then employed an algorithm to determine the modulation schemes \cite{dobre2007survey}. 

To remedy these problems, ML-based classifiers that aim to learn on preprocessed received data have been adopted and shown great advantages. ML algorithms usually provide better generalization to new unseen datasets, making their application preferable over solely FB approaches. For instance, the authors of \cite{wang2009recognitionAMR},  \cite{tabatabaei2010svm} and \cite{hassanpour2016automatic} used the support vector machine (SVM) machine learning algorithm to classify modulation schemes.
While, strictly FB approaches may become obsolete with the advent of the employment of ML classifiers for AMR, hand engineered features can provide useful input to ML techniques. For instance in the following works \cite{popoola2013effect} 
and \cite{jagannath2018artificial}, the authors engineered features using expert experience applied on the raw received signal and feeding the designed features as input for a neural network ML classifier.

Although ML methods have the advantage of better generality, classification efficiency and performance, the feature engineering step to some extent still depends on expert knowledge. As a consequence, the overall classification accuracy may suffer and depend on the expert input. On the other hand, current communication systems tend to become more complex and diverse, posing new challenges to the coexistence of homogeneous and heterogeneous signals and a heavy burden on the detection and recognition of signals in the complex radio environment.
Therefore, the ability of self-learning is becoming a necessity when confronted with such complex environment.

Recently, the wireless communication community experienced a breakthrough by adopting deep learning techniques to the wireless domain. In \cite{o2017learning}, deep convolution neural networks (CNNs) are applied directly on complex time domain signal data to classify modulation formats. The authors demonstrated that CNNs outperform expert-engineered features in combination with traditional ML classifiers, such as SVMs, k-Nearest Neighbors (k-NN), Decision Trees (DT), Neural Networks (NN) and Naive Bayes (NB).
An alternative method, is to learn the modulation format of the received signal from different representations of the raw signal. In our work in \cite{kulin2018end}, CNNs are employed to learn the modulation of various signals using the in-phase and quadrature (IQ) data representation  of the raw received signal and two additional data representations without affecting the simplicity of the input.
We showed that the amplitude/phase representation outperformed the other two, demonstrating the importance of the choice of the wireless data representation used as input to the deep learning technique so as to determine the most optimal 
mapping from the raw signal to the modulation scheme. 
Other, follow-up works include \cite{o2018over}, \cite{san2018evaluating}, \cite{mossaddeep2018}, \cite{rajendran2018deep},
\cite{tang2018digital}, \cite{meng2018automatic}, \cite{wu2018deep}, \cite{zhang2018automatic},
\cite{li2018generative}, \cite{li2018radioCNN}, \cite{yashashwi2019learnable},  \cite{sadeghi2019adversarial}, etc.

For a more comprehensive overview of the state-of-the art work on AMR we refer the reader to tables \ref{tab:intro:overviewAMR1} and \ref{tab:intro:overviewAMR2}. Table \ref{tab:intro:reviewtablesdescribtion} describes the structure used for tables \ref{tab:intro:overviewAMR1}, \ref{tab:intro:overviewAMR2}, \ref{tab:intro:survey_MAC} and	\ref{tab:intro:survey_network}.

\begin{table*}[tb]
	\centering
	\begin{adjustbox}{width=0.7\textwidth}
		\centering
		\begin{tabular}{p{5.5cm} p{7cm}}
			\toprule
			\textbf{Column name} & \textbf{Description}  \\
			\hline
			\toprule
			
			\textit{\textbf{Research Problem}} & The problem addressed in the work \\
			\textit{\textbf{Performance improvement}} & Performance improvement achieved in the work \\
			\textit{\textbf{Type of wireless network}} & The type of wireless networks considered in the work and/or for which the problem is solved \\
			\textit{\textbf{Data Type}} & Type of data used in the work, e.g. synthetic or real \\
			\textit{\textbf{Input Data}}& The data used as input for the developed machine learning algorithms \\
			\textit{\textbf{Learning Approach}} & Type of learning approach, e.g. traditional machine learning (ML) or deep learning (DL) \\
			\textit{\textbf{Learning Algorithm}} & List of learning algorithms used \\
			\textit{\textbf{Year}} & The year when the work was published \\
			\textit{\textbf{Reference}}& The reference to the analyzed work \\
			\bottomrule	
		\end{tabular}
	\end{adjustbox}
	\caption{Description of the structure for tables \ref{tab:intro:overviewAMR1}, \ref{tab:intro:overviewAMR2}, 
		\ref{tab:intro:survey_MAC} and	\ref{tab:intro:survey_network} }
	\label{tab:intro:reviewtablesdescribtion}
\end{table*}

\textbf{\textit{Wireless interference identification.}} WII essentially refers to identifying the type of wireless emitters (signal or technology) existing in the local radio environment, which can be immensely helpful information to investigate an effective interference avoidance and coexistence mechanisms. For instance, for technologies operating in the ISM bands in order to efficiently coexist it is crucial to know what type of other emitters are present in the environment (e.g. Wi-Fi, Zigbee, Bluetooth, etc.). 
Similar to AMR, FB and ML approaches (e.g. using time or frequency features) may be employed for technology recognition and signal identification approaches. Due to the development of deep learning applications for wireless signals classification, there has been significant success in applying it also for WII approaches.

For instance, the authors of \cite{selim2017spectrum} exploit the amplitude/phase difference representation to train a CNN model network to discriminate several radar signals from Wi-Fi and LTE transmissions. Their method was able to successfully recognize radar signals even under the presence of several interfering signals (i.e. LTE and Wi-Fi) at the same time, which is a key step for reliable spectrum monitoring.

In \cite{rajendran2018deep}, the authors make use of the average magnitude spectrum representation of the raw observed signal on a distributed architecture with low-cost spectrum sensors together with an LSTM deep learning classifier to discriminate between different wireless emitters, such as TETRA, DVB, RADAR, LTE, GSM and WFM. Results showed that their method is able to outperform conventional ML approaches and a CNN based architecture for the given task.

In \cite{schmidt2017wireless} the authors use the time domain quadrature (i.e. IQ) representation of the received signal and amplitude/phase vectors as input for CNN classifiers to learn the type of interfering technology present in the ISM spectrum. The results demonstrate that the proposed scheme is well suited for discriminating between Wi-Fi, ZigBee and Bluetooth signals. In \cite{kulin2018end}, we introduce a methodology for end-to-end learning from various signal representations and investigate also the frequency domain (FFT) representation of the ISM signals and demonstrate that the CNN classifier that used FFT data as input outperforms the CNN models used by the authors in \cite{schmidt2017wireless}.
Similarly, the authors of \cite{bitar2017wireless} developed a CNN model to facilitate the detection and identification of frequency domain signatures for 802.x standard compliant technologies. Compared to \cite{schmidt2017wireless} the authors in \cite{bitar2017wireless} make use of spectrum scans across the entire ISM region (80-MHz) and feed as input to a CNN model.

In \cite{maglogiannis2019enhancing} the authors used a CNN model to perform recognition of LTE and Wi-Fi transmissions based on two wireless signal representations, namely, the IQ and the frequency domain representation. The motivation behind this approach was to obtain accurate information about the technologies present in the local wireless environment so as to select an appropriate mLTE-U configuration that will allow fair coexistence with Wi-Fi in the unlicensed spectrum band.

Other examples include \cite{o2017spectral}, \cite{sun2018learning}, \cite{rajendran2019unsupervised}, \cite{yi2018interference},  etc.

In some applications like cognitive radio (CR) and spectrum sensing, the goal is however to identify the presence or absence of a signal. Namely, spectrum sensing is a process by which unlicensed users, also known as secondary users (SUs), acquire information about the status of the radio spectrum allocated to a licensed user, also known as primary user (PU), for the purpose of accessing unused licensed bands in an opportunistic manner without causing intolerable interference to the transmissions of the licensed user \cite{akyildiz2008survey}.

For instance, in \cite{soto2018detection} four ML techniques are examined k-NN, SVM, DT and logistic regression (LR) in order to predict the presence or absence of a PU in CR applications. The authors in \cite{lee2019deepsensing} go a step further and design a spectrum sensing framework based on CNNs to facilitate a SU to achieve higher sensing accuracy compared with conventional approaches. For more examples, we refer the reader to \cite{lee2018resource}, \cite{awe2018spatio} and \cite{han2017spectrum}.

The literature related to machine learning and deep learning for WII approaches is contained in Tables \ref{tab:intro:overviewAMR1} and \ref{tab:intro:overviewAMR2} ordered by the publishing year of the work. 

\begin{sidewaystable*}[ph!]
	\large
	\caption{An overview of work on machine learning for radio level analysis for performance optimization - PART 1}
	\label{tab:intro:overviewAMR1}
	\begin{adjustbox}{width=1\textwidth}
		\centering
		\begin{tabular}{p{3.8cm} p{7.5cm} p{4.8cm} p{2cm} p{7.0cm} p{2.2cm} p{6cm} p{1cm} c}
			\toprule
			\textbf{Research Problem} & \textbf{Performance improvement} & \textbf{Type of wireless network} & \textbf{Data Type}& \textbf{Input Data} & \textbf{Learning Approach} & \textbf{Learning Algorithm} & \textbf{Year} & \textbf{Reference} \\
			\hline
			\toprule
			AMR & More efficient spectrum utilization & Cognitive radio & Synthetic & FR, ZCR, RE  & ML & SVM & 2010 &  \cite{tabatabaei2010svm}\\
			\myrowcolour%
			AMR & More accurate signal modulation recognition for cognitive radio applications  & Cognitive radio & Synthetic & CWT, HOM & ML & ANN & 2010 & \cite{hassan2010automatic}\\
			
			Emitter identification & More accurate Radar Specific Emitter Identification  & Radar & Real & Cumulants & ML &k-NN & 2011 & \cite{aubry2011cumulants}\\
			\myrowcolour%
			AMR & More accurate signal modulation recognition for cognitive radio and DSA applications  & Cognitive radio & Synthetic & Max(PSD), NORM(A), AVG(x) & ML & ANN & 2011& \cite{popoola2011novel}\\
			
			AMR & More accurate signal modulation recognition for cognitive radio applications  & Cognitive radio & Synthetic & Cumulants & ML & k-NN & 2012 & \cite{aslam2012automatic}\\
			\myrowcolour%
			AMR & More efficient spectrum utilization & Cognitive radio & Synthetic & Max(PSD), STD(ap), STD(dp), STD(aa), STD(df), $F_c$, cumulants, CWT & ML & SVM & 2012 &  \cite{valipour2012automatic}\\
			
			AMR & More efficient spectrum utilization & Cognitive radio & Synthetic & $v_{20}$, AVG(A), $\beta$, Max(PSD),  STD(ap), STD(dp), STD(aa) & ML & FC-FFNN, FC-RNN & 2013 &  \cite{popoola2013effect}\\
			
			\myrowcolour%
			AMR & More accurate signal modulation recognition for cognitive radio and DSA applications  & Cognitive radio & Synthetic & ST, WT & ML & NN, SVM, LDA, NB, k-NN & 2015 & \cite{satija2015automatic}\\
			
			AMR & More efficient spectrum utilization & Cognitive radio & Synthetic & STD(dp), CWT, AVG(NORM())  & ML & SVM & 2016 &  \cite{hassanpour2016automatic}\\	
			\myrowcolour%
			AMR & More efficient spectrum utilization  & Cognitive radio & Real & IQ samples & DL \& ML & CNN, DNN, k-NN, DT, SVM, NB & 2016 &   \cite{o2016convolutional}\\
			
			AMR & More accurate signal modulation recognition for cognitive radio applications  & Cognitive radio & Synthetic & IQ samples & ML & DNN & 2016& \cite{kim2016deep}\\
			\myrowcolour%
			
			AMR & More accurate signal modulation recognition for cognitive radio applications  & Cognitive radio & Synthetic & IA signal samples, cumulants & DL \& ML& CNN, SVM & 2017& \cite{peng2017modulation}\\
			
			AMR & More accurate signal modulation recognition for cognitive radio applications  & Cognitive radio & Synthetic& Cumulants & DL & DNN, ANC, SAE & 2017& \cite{ali2017automatic}\\
			\myrowcolour%
			AMR & More accurate signal modulation recognition for cognitive radio applications  & Cognitive radio & Synthetic & IQ samples & DL & CLDNN, ResNet, DenseNet & 2017 & \cite{liu2017deep} \\
			
			AMR & More accurate signal modulation recognition for cognitive radio & Cognitive radio & Synthetic & IQ samples & DL & CNN, AE & 2017 & \cite{o2017introduction} \\
			\myrowcolour%
			Emitter identification & More efficient spectrum utilization  & Cognitive radio & Synthetic & IQ  samples & DL & AE, CNN & 2017 &  \cite{o2017semi}\\
			
			AMR & More efficient spectrum utilization  & Cognitive radio & Synthetic & IQ samples & DL & CLDNN, CNN, ResNet & 2017 &  \cite{west2017deep}\\
			\myrowcolour%
			TR & More efficient management of the wireless spectrum  & Cellular, WLAN, WPAN, WMAN & Real & Spectrograms & DL & CNN & 2017 &  \cite{o2017spectral}\\
			
			SI & More efficient spectrum utilization  & Cognitive radio & Synthetic & CFD, (non)standardized IQ  samples  &DL &CNN & 2017 &  \cite{han2017spectrum}\\
			\myrowcolour%
			AMR & More accurate and simple spectral events detection & ISM & Real & Spectograms & DL&YOLO & 2017 &  \cite{o2017learning}\\
			
			AMR & More efficient spectrum utilization  & Cognitive radio & Synthetic& IQ  samples & DL&CNN & 2017 &  \cite{karra2017modulation}\\			
			\myrowcolour%
			AMR & More efficient spectrum utilization  & Cognitive radio & Synthetic& IQ samples & DL&CNN & 2017 &  \cite{hauser2017signal}\\
			
			SI & More efficient spectrum monitoring  & Radar & Real& Spectrograms, A/Ph & DL&CNN & 2017 &  \cite{selim2017spectrum}\\
			\myrowcolour%
			WII & Improved spectrum utilization & Bluetooth, Zigbee, Wi-Fi & Real& Power-frequency & DL&CNN & 2017 &  \cite{bitar2017wireless}\\
			
			SI & More efficient spectrum monitoring  & Cognitive radio & Real& FFT & DL&CNN & 2017 &  \cite{paisana2017context}\\
			\myrowcolour%
			WII &  More efficient spectrum management via wireless interference identification & Bluetooth, Zigbee, Wi-Fi & Synthetic& FFT  & DL &CNN, NFSC & 2017 &  \cite{schmidt2017wireless}\\
			
			Emitter identification & More accurate Radar Specific Emitter Identification  & Radar & Real \& Synthetic& FD-curves & ML &SVM & 2018 &  \cite{zhao2018specific}\\			
			\myrowcolour%
			AMR & More efficient spectrum utilization  & Cognitive radio & Real & In-band spectral variation, deviation from unit circle, cumulants & ML & ANN, HH-AMC & 2018 &  \cite{jagannath2018artificial}\\
			
			\bottomrule
		\end{tabular}
	\end{adjustbox}
\end{sidewaystable*}


\begin{sidewaystable*}[ph!]
	\large
	\caption{An overview of work on machine learning for radio level analysis for performance optimization - PART 2}
	\label{tab:intro:overviewAMR2}
	\begin{adjustbox}{width=1\textwidth}	
		\centering
		\begin{tabular}{p{3.8cm} p{7.5cm} p{4.8cm} p{2cm} p{7.0cm} p{2.2cm} p{6cm} p{1cm} c}
			\toprule
			\textbf{Research Problem} & \textbf{Performance improvement} & \textbf{Type of wireless network} & \textbf{Data Type}& \textbf{Input Data} & \textbf{Learning Approach} & \textbf{Learning Algorithm} & \textbf{Year} & \textbf{Reference} \\
			\hline
			\toprule
			\myrowcolour%
			AMR & More efficient spectrum utilization  & Cognitive radio & Synthetic & IQ samples, HOMs & DL& CNN, RN & 2018 &  \cite{o2018over}\\
			
			AMR & More accurate signal modulation recognition for cognitive radio and DSA applications  & Cognitive radio & Synthetic & WT, CFD, HOMs, HOCs, ST & SVM & 2018 & \cite{ghasemzadeh2018performance}\\
			
			
			\myrowcolour%
			AMR & Modulation and Coding Scheme recognition for cognitive radio and DSA applications & Wi-Fi & Synthetic & IQ  samples & DL& CNN & 2018& \cite{san2018evaluating} \\
			
			AMR & More accurate signal modulation recognition for cognitive radio and DSA applications  & Cognitive radio & Synthetic & IQ  samples, FFT & DL& CNN, CLDNN, MTL-CNN & 2018 & \cite{mossaddeep2018} \\
			\myrowcolour%
			AMR & More efficient spectrum utilization  & Cognitive radio & Synthetic & A/Ph, $AVG_mag_FFT$ & DL \& ML & CNN, LSTM, RF, SVM, k-NN & 2018 &  \cite{rajendran2018deep}\\

			TR &  More efficient spectrum management via wireless interference identification & Bluetooth, Zigbee, Wi-Fi & Synthetic & IQ samples & DL & CNN & 2018 &  \cite{grunau2018multi}\\
			\myrowcolour%
			WII &  More efficient spectrum utilization by detecting interference & Bluetooth, Zigbee, Wi-Fi & Real & RSSI samples & DL & CNN & 2018 &  \cite{yi2018interference}\\
			
			AMR & More efficient spectrum utilization  & Cognitive radio & Synthetic & Contour Stellar Image & DL& CNN,  AlexNet, ACGAN & 2018 &  \cite{tang2018digital}\\
			\myrowcolour%
			AMR & More efficient spectrum utilization   & Cognitive radio & Synthetic & Constellation diagram & DL & CNN, AlexNet, GoogLeNet & 2018 &  \cite{peng2018modulation}\\
			
			AMR & More efficient spectrum utilization & UAV & Synthetic& IQ  samples & CNN, LSTM & 2018 &  \cite{zhang2018automaticuav}\\
			\myrowcolour%
			SI & More efficient spectrum utilization   & Cognitive radio & Synthetic & Activity, non-activity, "amplitude node" and permanence probability  & ML & k-NN, SVM, LR, DT & 2018 &  \cite{soto2018detection}\\
			
			AMR & More efficient spectrum utilization & Cognitive radio & Synthetic & IQ samples & DL& CNN & 2018 &  \cite{meng2018automatic}\\
			\myrowcolour%
			AMR &  More efficient spectrum utilization by detecting interference & RF signals & Synthetic & IQ samples & DL \& ML& SVM, DNN, CNN, MST & 2018 &  \cite{youssef2018machine}\\
			
			AMR & More efficient spectrum utilization & Cognitive radio & Synthetic & IQ samples  & DL \& ML & CNN, LSTM, SVM & 2018 &  \cite{wu2018deep}\\
			\myrowcolour%
			AMR & More efficient spectrum utilization & VHF & Synthetic & IQ samples  & DL& CNN & 2018 &  \cite{wu2018vhf}\\
			
			Emitter identification & More efficient spectrum utilization & Cognitive radio & Synthetic & IQ signal samples, FOC  & DL& CNN, LSTM & 2018 &  \cite{zhang2018automatic}\\
			\myrowcolour%
			SI & More efficient spectrum utilization & Cellular & Synthetic& IQ and Amplitude data  & DL& CNN & 2018 &  \cite{duan2018automatic}\\
			
			AMR & More efficient spectrum utilization & Cognitive radio & Synthetic& IQ samples  & DL \& ML &ACGAN, SCGAN, SVM, CNN, SSTM & 2018 &  \cite{li2018generative}\\
			\myrowcolour%
			SI & Improved spectrum sensing & Cognitive radio & Synthetic & Beamformed IQ samples  & ML& SVM & 2018 &  \cite{awe2018spatio}\\
			
			AMR & More efficient spectrum utilization & Cognitive radio & Synthetic & IQ samples  & DL \& ML& DCGAN, NB, SVM, CNN & 2018 &  \cite{li2018radioCNN}\\
			\myrowcolour%
			
			AMR & More efficient spectrum utilization & Cognitive radio & Synthetic & IQ &DL& RNN & 2018 &  \cite{hu2018robust}\\	
			
			AMR & More efficient spectrum utilization & Cognitive radio & Synthetic & IQ samples with corrected frequency and phase
			&DL& CNN, CLDNN & 2019 &  \cite{yashashwi2019learnable}\\
			\myrowcolour%
			TR & More efficient management of the wireless spectrum  & LTE and Wi-Fi & Real& IQ samples, FFT &DL& CNN & 2019 &  \cite{maglogiannis2019enhancing}\\
			
			AMR & More efficient spectrum utilization   & Cognitive radio & Synthetic & IQ samples &DL& CNN & 2019 &  \cite{sadeghi2019adversarial}\\		
			\myrowcolour%
			SI & Improved spectrum sensing & Cognitive radio & Synthetic& RSSI  &DL& CNN & 2019 &  \cite{lee2019deepsensing}\\
			
			WII &  More efficient management of the wireless spectrum  & Bluetooth, Zigbee, Wi-Fi & Real& FFT, A/Ph &DL& CNN, LSTM, ResNet, CLDNN & 2019 &  \cite{zhang2019deep}\\
			\myrowcolour%
			WII &  More efficient management of the wireless spectrum  & Sigfox, LoRA and IEEE 802.15. 4g & Real& IQ, FFT &DL& CNN & 2019 &  \cite{shahid2019convolutional}\\
			
			WII & More accurate and simple spectral events detection  & Cognitive radio & Synthetic \& Real & PSD data & DL &AAE & 2019 &  \cite{rajendran2019unsupervised}\\
			\myrowcolour%
			TR & More efficient management of the wireless spectrum  & GSM, WCDMA and LTE & Synthetic & SCF, FFT, ACF, PSD &DL& SVM & 2019 &  \cite{tekbiyik2019multi}\\
			
			TR & More efficient management of the wireless spectrum  & LTE, Wi-Fi and DVB-T & Real & RSSI, IQ and Spectogram &DL& CNN & 2019 &  \cite{fontaine2019towards}\\
			\myrowcolour%
			AMR & More efficient spectrum utilization & Cognitive radio & Synthetic & IQ, A/Ph &DL& CLDNN, ResNet, LSTM & 2019 &  \cite{sadeghi2019adversarial}\\	
			
			\bottomrule
		\end{tabular}%
	\end{adjustbox}
\end{sidewaystable*}


\subsubsection{Medium access control (MAC) analysis}
Sharing the limited spectrum resources is the main concern in wireless networks \cite{isolani2018survey}.
One of the key functionalities of the MAC layer in wireless networks is to negotiate the access to the wireless medium to share the limited resources in an ad hoc manner. Opposed to centralized designs where entities like base
stations control and distribute resources, nodes in Wireless Ad hoc Networks (WANETs) have to coordinate resources.

For this purpose, several MAC protocols have been proposed in the literature. Traditional MAC protocols designed for WANETs include Time Division Multiple Access (TDMA) \cite{cordeiro2007c}, \cite{hadded2015tdma}, Carrier Sense Multiple Access/Collision Avoidance (CSMA/CA) \cite{lien2008carrier}, \cite{jain2001multichannel},  Code Division Multiple Access (CDMA) \cite{muqattash2003cdma}, \cite{kumar2006medium} and hybrid approaches \cite{sitanayah2010er}, \cite{su2007opportunistic}.
However, given the changing network and environment conditions, designing a MAC protocol that fits all possible conditions and various application requirements is a challenge especially when these conditions are not available
or known a priori.
This subsection investigates the advances made related to the MAC layer to tackle the problem of efficient spectrum sharing  with the help of machine learning. We identify two categories of MAC analysis i) MAC identification  and ii) Interference recognition.
The reviewed MAC analysis tasks are listed in Table \ref{tab:intro:survey_MAC}.

\textbf{\textit{MAC identification.}} These approaches are typically employed in cognitive radio (CR) applications to foster communication and coexistence between protocol-distinct technologies. CRs rely on information gathered during spectral sensing to infer the environment conditions, presence of other technologies and spectrum holes.
Spectrum holes are frequency bands that have been allocated to licensed network users but are not used at a particular time, which can be utilized by a CR user. Usually, spectrum sensing can determine the frequency range of a spectrum hole,
while the timing information, which is also a channel access parameter, is unknown.

MAC protocol identification approaches may help CR users determine the timing information of a spectrum hole and accordingly tailor its packet transmission duration, which provides the potential benefits for network performance improvement.
For this purpose several MAC layer characteristics can be exploited.

For example, in \cite{yang2010mac} the TDMA and slotted ALOHA MAC protocols are identified based on two features, the power mean and the power variance of the received signal combined with a SVM classifier.
The authors in \cite{hu2014mac} utilized power and time features to distinguish between four MAC protocols, namely TDMA, CSMA/CA, pure ALOHA, and slotted ALOHA using a SVM classifier. 
Similary, in \cite{rajab2015energy} the authors captured MAC layer temporal features of 802.11 b/g/n homogeneous and heterogeneous networks and employed a k-NN and a NB ML classifier to distinguish between all three. 

\textbf{\textit{Interference recognition.}}
Similar to the approaches of recognizing interference based on radio spectrum analysis, the goal here is to identify the 
type of radio interference which degrades the network performance.
However, compared to the previously introduced work, the works in the MAC analysis level category focus on identifying distinct features of interfered channel and packets to detect and quantify interference in order to pinpoint the viability of opportunistic transmissions in interfered channels and select an appropriate strategy to co-exist with the present interference.
This is realized based on information available on low-cost off-the-shelf devices, such as 802.15.4 and Wi-Fi radios, which is used as input for ML classifiers.

For instance, in \cite{hithnawi2015tiim} the authors investigated two possibilities for detecting interference: i) the energy variations during packet reception captured by sampling the radio's RSSI register and ii) monitoring the Link Quality Indicator (LQI) of received corrupted packets. This information is combined with a DT classifier, considered as a computationally and memory efficient candidate for the implementation on 802.15.4 devices.
Another work on interference identification in WSNs is \cite{hermans2013sonic}. The authors were able to accurately distinguish Wi-Fi, Bluetooth and microwave oven interference based on features of corrupted packets (i.e. mean normalized RSSI, LQI, RSSI range, error burst spanning and mean error burst spacing) used as input to a SVM and DT classifier.

In \cite{rayanchu2011airshark} the authors were able to detect non-Wi-Fi interference on Wi-Fi commodity hardware. 
They collected energy samples across the spectrum from the Wi-Fi card to extract a diverse set of features  that capture the
spectral and temporal properties of wireless signals (e.g. central frequency, bandwidth, spectral signature, duty cycle, pulse signature, inter-pulse timing signature, etc.). They used these features and investigated performance of two classifiers, SVM and DT. The idea is to embed these functionalities in Wi-Fi APs and clients, which can then implement an appropriate mitigation mechanisms that can quickly react to the presence of significant non Wi-Fi interference. 

\begin{sidewaystable*}[ph!]
	\large
	\caption{An overview of work on machine learning for MAC level analysis for performance optimization}
	\label{tab:intro:survey_MAC}
	\begin{adjustbox}{width=1\textwidth}
		\centering
		\begin{tabular}{p{3.8cm} p{7.5cm} p{4.8cm} p{2cm} p{8.0cm} p{2.2cm} p{5cm} p{1cm} c}
			\toprule
			\textbf{Research Problem} & \textbf{Performance improvement} & \textbf{Type of wireless network} & \textbf{Data Type}& \textbf{Input Data} & \textbf{Learning Approach} & \textbf{Learning Algorithm} & \textbf{Year} & \textbf{Reference} \\			
			\hline
			\toprule   
			
			MAC identification & More efficient spectrum utilization & Cognitive radio & Synthetic& Mean and variance of power samples  & ML & SVM & 2010 &  \cite{yang2010mac}\\
			\myrowcolour%
			Wireless interference identification &  Enhanced spectral efficiency by interference mitigation & Wi-Fi & Real& Duty cycle, Spectral signatures, Frequency and bandwidth, Pulse signatures, Pulse spread, Inter-pulse timing signatures, Frequency sweep & ML& DT&  2011 &  \cite{rayanchu2011airshark}\\
			
			MAC identification & More efficient spectrum utilization & Cognitive radio & Synthetic&Power mean, Power variance, Maximum power, Channel busy duration, Channel idle duration  & ML&SVM, NN, DT & 2012 &  \cite{hu2012mac}\\ 
			\myrowcolour%
			Wireless interference identification &  Enhanced spectral efficiency by interference mitigation & Wi-Fi, Bluetooth, Microwave for WSN& Real& LQI, range(RSSI), Mean error burst spacing, Error burst spanning, AVG(NORM(RSSI)), 1 - mode(RSSI$_{normed}$) & ML&SVM, DT & 2013 &  \cite{hermans2013sonic}\\						
			
			MAC identification & More efficient spectrum utilization & Cognitive radio & Synthetic& Received power mean, Power variance, Channel busy state duration, Channel idle state duration  &ML&SVM & 2014 &  \cite{hu2014mac}\\ 
			\myrowcolour%
			Wireless interference identification &  Reduced power consumption by interference identification & Wireless sensor network & Real& On-air time, Minimum Packet Interval,  Peak to Average Power Ratio, Under Noise Floor & ML&DT & 2014 &  \cite{zheng2014zisense}\\
			
			MAC identification & More efficient spectrum utilization & WLAN & Real&Number of activity fragments at 111$\mu s$, between 150$\mu s$ and  200$\mu s$, between 200$\mu s$	and 300$\mu s$, between 300$\mu s$ and 500$\mu s$, and number of fragments between 1100$\mu s$ and 1300$\mu s$  & ML& k-NN, NB & 2015 &  \cite{rajab2015energy}\\
			\myrowcolour%
			Wireless interference identification &  Enhanced spectral efficiency by interference mitigation & Wireless sensor network & Real&
			Packet corruption rate, Packet loss rate, Packet length, Error rate, 
			Error burstiness, Energy perception per packet,  Energy perception level per packet,
			Backoffs,  Occupancy level, Duty cycle,  Energy span during packet reception,
			Energy level during packet reception, RSSI regularity during packet reception & ML &DT & 2015 &  \cite{hithnawi2015tiim}\\
			
			Spectrum prediction & Enhanced performance via more efficient spectrum usage achieved with medium availability prediction & ISM technologies & Synthetic& State matrix of the network, $X^{f,n}$ for each node $n$ in frame $f$ & ML & NN& 2018 &  \cite{mennes2018neural}\\
			\myrowcolour%
			Spectrum prediction & Enhanced performance via more efficient spectrum usage achieved with select a predicted free channel & WSN & Synthetic and Real &DNN with network states as input and Q values as output & DL &DQN & 2018 &  \cite{wang2018deep}\\ 
			
			Spectrum prediction & More efficient radio resource utilization and enhanced link scheduling and power control & Cellular & Synthetic& The channel	matrix \boldmath$H$ containing $|h_{i,j}|^2$ between all pairs of transmitter
			and receivers and the weight matrix \boldmath$W$  & DL & DQN and DNN & 2018 &  \cite{xu2018realtime}\\
			\myrowcolour%
			Spectrum prediction & More efficient spectrum utilization and increased network performance & CRN & Synthetic&Environmental states $s$. & DL & ResNet, DNN, DQN& 2019 &  \cite{yu2019deep}\\	
			
			Spectrum prediction & More efficient spectrum utilization and increased network performance via predicting PUs future activity & CRN & Synthetic&Vector \boldmath$x$ with $n$ channel sensing results, where each result has value "-1" (idle) or "1" (busy).& ML & NN& 2019 &  \cite{zhang2019neural}\\	
			\myrowcolour%
			Spectrum prediction & Enhanced performance via more efficient spectrum usage achieved with medium availability prediction & ISM technologies & Synthetic and Real&Channel observations $O_{s,c}^{f,n}$ on channel $c$ made by node $n$ at time $(f, s)$, where $s$ is a timeslot in superframe $f$ & DL &CNN & 2019 &  \cite{mennes2019deep}\\
			
			\bottomrule
		\end{tabular}%
	\end{adjustbox}
\end{sidewaystable*}

The authors of \cite{zheng2014zisense} propose an energy efficient rendezvous mechanism resilient to interference for WSNs based on ML. Namely, due to the energy constraints on sensor nodes, it is of great importance to save energy and extend the network lifetime in WSNs. Traditional rendezvous mechanism such as Low Power Listening (LPL) and Low Power Probe (LPP) rely on low duty cycling (scheduling the radio of a sensor node between ON and OFF compared to always-ON methods) depending on the presence of a signal (e.g. signal strength). However, both suffer performance degradation in noisy environments with signal interference incorrectly regarding a non-ZigBee interfering signal as an interested signal and improperly keeping the radio ON, which increases the probability of false wake-ups. To remedy this, the proposed approach in \cite{zheng2014zisense} is capable of detecting potential ZigBee transmissions and accordingly decide whether to turn the radio ON. For this purpose, they extracted signal features from time domain RSSI samples (i.e. On-air time, Minimum Packet Interval, Peak to Average Power Ratio and Under Noise Floor) and used it as input to a DT classifier to effectively distinguish ZigBee signals from other interfering ones.

\textbf{\textit{Spectrum prediction.}}
In order to share the available spectrum in a more efficient way, there are various attempts in predicting the wireless medium availability to minimize transmission collisions and, therefore, increase the overall performance of the network.

For instance, an intelligent wireless device may monitor the medium and based on MAC-level measurements predict if the medium is likely to be busy or idle. In another variation of this approach, a device may predict the quality of the channels in terms of properties such as idle probabilities or idle durations and then select the channel with the highest quality for transmission.

For instance, the authors in \cite{mennes2018neural} use NNs to predict if a slot will be free based on some history to minimize collisions and optimize the usage of the scarce spectrum. In their follow up work \cite{mennes2019deep}, they exploit CNNs to predict the spectrum usage of the other neighboring networks. Their approach is aimed for devices with limited capabilities for retraining.

In \cite{wang2018deep}, a Deep Q-Network (DQN) is proposed to predict and select a free channel for WSNs.
In \cite{zhang2019neural}, the authors design a NN predictor to predict PUs future activity based on past channel occupancy sensing results, with the goal of improving secondary users (SUs) throughput while alleviating collision to primary user (PU) in full-duplex (FD) cognitive networks.	 

The authors of \cite{yu2019deep} consider the problem of sharing time slots among a multiple of time-slotted networks so as to maximize the 
sum throughput of all the networks. The authors utilize the ResNet and compare performance to a plain DNN.

MAC analysis approaches are listed in Table \ref{tab:intro:survey_MAC}.

\subsubsection{Network prediction}
Network prediction  refers to tasks related to inferring the wireless network performance or network traffic, given historical measurements or related data. 
Table \ref{tab:intro:survey_network} gives an overview of the works on machine learning for network level prediction tasks, i.e. i) Network performance prediction and ii) Network traffic prediction.

\textbf{\textit{Network performance prediction}}. 
ML approaches are used, extensively, to create prediction models for many wireless networking applications. Typically, the goal is to forecast the performance or optimal device parameters/settings and use this knowledge to adapt the communication parameters to the changing environment conditions and application QoS requirements so as to optimize the overall network performance.

For instance, in \cite{al2017new} the authors aim  to select the optimal MAC parameter settings in 6LoWPAN networks to reduce excessive collisions, packet losses and latency. First, the MAC layer parameters are used as input to a NN to predict the throughput and latency, followed by an optimization algorithm to achieve high throughput with minimum delay.
The authors of \cite{pierucci2016neural} employ NNs to predict the users’ QoE in cellular networks, based on average user throughput, number of active users in a cells, average data volume per user and channel quality indicators, demonstrating high prediction accuracy.

Given the dynamic nature of wireless communications, a traditional one-MAC-fit-all approach cannot meet the challenges under significant dynamics in operating conditions, network traffic and application requirements. 
The MAC protocol may deteriorate significantly in performance as the network load becomes heavier, while the protocol may waste network resources when the network load turns lighter. 
To remedy this, \cite{sha2013self} and \cite{qiao2016mac} study an adaptive MAC layer with multiple MACs available that is able to select the MAC protocol most suitable for the current conditions and application requirements.
In \cite{sha2013self} a MAC selection engine for WSNs based on a DT model decides which is the best MAC protocol for the given application QoS requirements, current traffic pattern and ambient interference levels as input. The candidate protocols are TDMA, BoX-MAC and RI-MAC.
The authors of \cite{qiao2016mac} compare the accuracy of NB, Random Forest (RF), decision trees and SMO \cite{keerthi2001improvements} to decide between the DCF and TDMA protocol to best respond to the dynamic network circumstances.

In \cite{baldo2009neural} an NN model is employed that learns how environmental measurements and the status of the network affect the performance experienced on different channels, and uses this knowledge to dynamically select the channel which is expected to yield the best performance for the user.

As an integral part of reliable communication in WSNs, accurate link estimation is essential for routing
protocols, which is a challenging task due to the dynamic nature of wireless channels. To address this problem,
the authors in \cite{liu2011foresee} use ML (i.e. LR, NB and  NN) to predict the link quality based on physical layer parameters of last received packets and the PRR, demonstrating high accuracy and improved routing.
The same authors go a step further in \cite{Liu2014temporal} and employ online machine learning to
adapt their link quality prediction mechanism real-time to the notoriously dynamic wireless environment.

The authors in \cite{kulin2017poster} develop a ML engine that predicts the packet loss rate in a WSN using machine learning techniques network performance as an integral part for an adaptive MAC layer.

\begin{sidewaystable*}[ph!]
	\large
	\caption{An overview of work on machine learning for network level analysis for performance optimization}
	\label{tab:intro:survey_network}
	\begin{adjustbox}{width=1\textwidth}
		\centering
		\begin{tabular}{p{3.8cm} p{7.5cm} p{4.8cm} p{2cm} p{8.0cm} p{2.2cm} p{5cm} p{1cm} c}
			\toprule
			\textbf{Research Problem} & \textbf{Performance improvement} & \textbf{Type of wireless network} & \textbf{Data Type}& \textbf{Input Data} & \textbf{Learning Approach} & \textbf{Learning Algorithm} & \textbf{Year} & \textbf{Reference} \\			
			\hline
			\toprule   
			Network performance prediction &  Enhanced resource utilization by predicting the network state & Wi-Fi & Synthetic&Packet Rate, Data Rate, CRC Error Rate, PHY Error Rate, Packet
			Size, Day fo Week, Hour of Day  & ML& MFNN & 2009 &  \cite{baldo2009neural}\\
			
			Network performance prediction &  Enhanced performance by accurate link quality prediction & Wireless sensor network & Real& PRR, RSSI, SNR and LQI & ML& LR, NB and  NN & 2011 &  \cite{liu2011foresee}\\
			\myrowcolour%
			Network performance prediction &  Enhanced performance by accurate link quality prediction & Wireless sensor network & Real&PRR, RSSI, SNR and LQI  & ML& LR & 2012 &  \cite{liu2012talent}\\
			
			Network performance prediction &  Enhanced performance by selecting the optimal MAC scheme & Wireless sensor network & Real& RSSI statistics (mean and variance), IPI statistics (mean and variance), reliability, energy consumption and latency & ML& DT & 2013 &  \cite{sha2013self}\\
			\myrowcolour%
			Network performance prediction &  Enhanced performance by accurate link quality prediction & Wireless sensor network & Real&PRR, RSSI, SNR and LQI  &ML&LR, SGD & 2014 &  \cite{Liu2014temporal}\\
			
			Network performance prediction &  Enhanced network performance via performance characterization and  optimal radio parameters prediction& Cellular & Synthetic&SINR (Signal to Interference plus Noise Ratio), ICI, MCS, Transmit power  & ML& Random NN & 2015 &  \cite{adeel2015critical}\\  
			\myrowcolour%
			Network traffic prediction &  Enhanced resource allocation by predicting mobile traffic demand & Cellular & Real&Traffic (Bytes)per 10min & ML&Hierarchical clustering & 2015 &  \cite{wang2015understanding1}\\	
			
			Network traffic prediction &  Enhanced base station sleeping mechanism which reduced power consumption by predicting mobile traffic demand & Cellular & Synthetic & Machine learning & 2015 &  \cite{hu2015base}\\	
			\myrowcolour%
			Network traffic prediction &  Enhanced resource allocation by predicting mobile traffic demand & Cellular & Real& User traffic & ML&MWNN & 2015 &  \cite{zang2015wavelet}\\
			
			Network performance prediction &  Enhanced QoE by predicting KPI parameters & Cellular & Real& Mobile network KPIs& ML& NN & 2016 &  \cite{pierucci2016neural}\\
			\myrowcolour%
			Network performance prediction & Enhancing performance by selecting the optimal MAC & Cognitive radio & Synthetic& Protocol Type, Packet Length, Data Rate , Inter-arrival Time , Transmit Power, 
			Node Number, Average load, Average throughput, Transmitting delay, Minimum throughput, Maximum throughput, Throughput standard deviation and classification result & ML& NB, DT, RF, SMO & 2016 &  \cite{qiao2016mac}\\		
			
			Network traffic prediction &  Enhanced resource allocation by predicting mobile traffic demand & Cellular & Real&Mobile traffic volume & ML&Regression analysis & 2016 &  \cite{xu2016bigts}\\
			\myrowcolour%
			Network traffic prediction &  Enhanced resource allocation by predicting mobile traffic demand & Cellular & Real& Mobile traffic & ML& SVM, MLPWD, MLP & 2016 &  \cite{nikravesh2016mobile}\\
			
			Network performance prediction &  Enhanced performance by predicting packet loss rate & WSN & Real&Number of detected nodes, IPI, Number of received packets, Number of erroneous packets
			LR, RT, NN & ML& & 2017 &  \cite{kulin2017poster}\\ 
			
			Network traffic prediction &  Enhanced resource allocation by predicting mobile traffic demand & Cellular & Real&Average traffic load per hour & DL & LSTM, GSAE, LSAE & 2017 &  \cite{wang2017spatiotemporal}\\
			\myrowcolour%
			Network traffic prediction &  Enhanced resource allocation by predicting mobile traffic demand & Cellular & Real&Number of CDRs generated during each time interval in a square of the Milan Grid & DL& RNN, 3D CNN & 2017 &  \cite{huang2017study}\\
			
			Network performance prediction &  Maximized reliability and minimized end-to-end delay by selecting optimal MAC parameters & 6LoWPAN & Synthetic&
			Maximum CSMA backoff, Backoff exponent, Maximum frame retries limit & ML&ANN & 2017 &  \cite{al2017new}\\ 
			\myrowcolour%
			Network traffic prediction & Enhanced resource allocation by predicting mobile traffic demand & Cellular & Real&Traffic volume snapshots every 10 min & DL& STN, LSTM, 3D CNN & 2018 &  \cite{zhang2018long}\\ 
			
			Network traffic prediction &  Enhanced resource allocation by predicting mobile traffic demand & Cellular & Real& Cellular traffic load per half-hour & DL&LSTM,GNN & 2018 &  \cite{wang2018spatio}\\
			\myrowcolour%
			Network traffic prediction &  Enhanced resource allocation by predicting mobile traffic demand & Cellular & Real&CDRs with an interval of 10 minutes & DL&DNN, LSTM & 2018 &  \cite{alawe2018improving}\\
			
			Network performance prediction &  Enhanced network resource allocation by predicting network parameters & Wireless sensor network & Synthetic&Network lifetime, Power level, Internode distance & ML& NN& 2018 & \cite{akbas2018neural}\\ 
			\myrowcolour%
			Network traffic prediction &  Enhanced resource allocation by predicting mobile traffic demand & Cellular & Real&SMS and Call volume per 10min interval & DL & CNN & 2018 &  \cite{zhang2018citywide}\\
			
			Network traffic prediction &  Enhanced resource allocation by predicting mobile traffic demand & Cellular & Real&Traffic load per 10 min & DL&LSTM & 2018 &  \cite{feng2018deeptp}\\
			\myrowcolour%
			Network traffic prediction & Enhanced resource allocation by more efficiently predicting mobile traffic load  & Cellular & Real &Traffic logs recorded at 10-minutes intervals & ML& RF & 2018 &  \cite{yamada2018feature}\\
			
			Network traffic prediction &  Enhanced resource allocation by predicting mobile traffic demand & Cellular & Real&Traffic logs recorded at 15-minutes intervals  & DL& LSTM & 2019 &  \cite{hua2019traffic}\\
			\myrowcolour%
			Network traffic prediction &  Enhanced resource allocation by predicting mobile traffic demand & Cellular & Real& Traffic load per 5-minute intervals  & DL& 3D CNN & 2019 &  \cite{bega2019deepcog}\\
			
			Network traffic prediction &  Enhanced resource allocation by predicting idle time windows & Cellular & Real&Number of unique subscribers observed and Number of communication events occurring in a counting time window for a specific cell & DL&TGCN, TCN, LSTM, and GCLSTM & 2019 &  \cite{fang2019idle}\\
			
			\bottomrule
		\end{tabular}%
	\end{adjustbox}
\end{sidewaystable*}

\textbf{\textit{Network traffic prediction}}. 
Accurate prediction of user traffic in cellular networks is crucial to evaluate and improve the system performance.
For instance, the functional base station sleeping mechanism may be adapted by utilizing knowledge about future
traffic demands, which are in  \cite{hu2015base} predicted based on a NN model. This knowledge helped reduce the overall power consumption, which is becoming an important topic with the growth of the cellular industry.

In another example, consider the need for efficient management of expensive mobile network resources, such as spectrum,
where finding a way to predict future network use can help for network resource management and planning. A new paradigm for future 5G networks is \textit{network slicing} enabling the network infrastructure to be divided into slices devoted to different services and tailored to their needs \cite{zhang2017network}. With this paradigm, it is essential to allocate the needed resources to each slice, which requires the ability to forecast their respective demands.	The authors in \cite{bega2019deepcog} employed a CNN model that, based on traffic observed at base stations of a particular network slice, predicts the capacity required to accommodate the future traffic demands for services associated to it.

In \cite{wang2017spatiotemporal} LSTMs are used to model the temporal correlations of the mobile traffic distribution and perform forecasting together with stacked Auto Encoders for spatial feature extraction. Experiments with a real-world dataset demonstrate superior performance over SVM and the Autoregressive Integrated Moving Average (ARIMA) model.

Deep learning was also employed in \cite{huang2017study}, \cite{feng2018deeptp} and \cite{alawe2018improving} where the authors utilize CNNs and LSTMs to perform mobile traffic forecasting. By effectively extracting spatio-temporal features, their proposals gain significantly higher accuracy than traditional approaches, such as ARIMA. 

Forecasting with high accuracy the volume of data traffic that mobile users will consume is becoming increasingly important for demand-aware network resource allocation. More example approaches can be found in \cite{hua2019traffic}, \cite{nikravesh2016mobile}, \cite{wang2017spatiotemporal}, \cite{zhang2018long}, \cite{wang2018spatio}, \cite{zhang2018citywide}, \cite{yamada2018feature}, \cite{bega2019deepcog} and \cite{fang2019idle}.

\subsection{\change{Machine Learning} applications for Information processing}
Wireless sensor nodes and mobile applications installed on various mobile devices record
application level data frequently, making them act as sensor hubs responsible for data acquisition
and preprocessing and subsequently storing the data in the "cloud" for further
"offline" data storage and real-time computing using big data technologies (e.g. Storm \cite{iqbal2015big}, Spark \cite{zaharia2016apache}, Kafka \cite{ranjan2014streaming}, Hadoop \cite{dittrich2012efficient}, etc).
Example applications are i) \textbf{\textit{IoT infrastructure monitoring}} such as smart farming \cite{lottes2017uav, sa2018weednet}, smart mobility \cite{nguyen2016traffic, rathore2015efficient}, smart city \cite{strohbach2015towards, amato2017deep, mittal2016spotgarbage} and smart grid \cite{gillis2016nonintrusive}, ii) \textbf{\textit{device fingerprinting}}, iii) \textbf{\textit{localization}} and iv) \textbf{\textit{activity recognition}}.

For instance, the works \cite{lv2017device}, \cite{jafari2018iot}, \cite{merchant2018deep}, 
\cite{thing2017ieee},  \cite{uluagac2013passive}, \cite{bezawada2018behavioral}, \cite{riyaz2018deep}
exploit various time and radio patterns of the data with machine learning classifiers to distinguish legitimate wireless devices from adversarial ones, so as to increase the wireless network security.

In the works of \cite{wang2015phasefi}, \cite{wang2016csi}, \cite{wang2017csi}, \cite{wang2015deepfi}, \cite{wang2017device} and \cite{zhang2016deeplearning} ML or deep learning is  employed to localize users
in indoor or outdoor environments, based on different signals received from wireless devices or about the wireless channels such as amplitude and phase channel state information (CSI), RSSI, etc.

The goal in the works \cite{zeng2016wiwho}, \cite{zhao2018through}, \cite{lv2019robust}, \cite{shahzad2018augmenting}, \cite{wang2015understanding} is to identify the activity of a person based on various wireless signal properties in combination with a machine learning technique. For instance, in \cite{zhao2018through} the authors demonstrate accurate human pose estimation through walls and occlusions based on properties of Wi-Fi wireless signals and how they reflect off the human body, used as input to a CNN classifier. In \cite{lv2019robust} the authors detect intruders based on how their movement patterns affect Wi-Fi signals in combination with a Gaussian Mixture Model (GMM).

For a more throughout overview of the applications and works on wireless information processing the reader is referred to \cite{mahdavinejad2018machine}.

\section{Open Challenges and Future Directions}
\label{sec:challenges}
Previous sections presented the significant amount of research work focused on exploiting ML to address the spectrum scarcity problem in future wireless networks.
However, despite the growing state-of-the-art with more and more different ML algorithms being explored and applied at various layers of the network protocol stack, there are still open challenges that need to be addressed in order to employ these paradigms in real radio environments to enable a fully intelligent wireless network in the near future. 

This section discusses a set of open challenges and explores future research directions which are expected to accelerate the adoption of ML in future wireless network deployments.

\subsection{Standard Datasets, Problems, Data representation and \change{Evaluation metrics}}

\subsubsection{Standard Datasets}
To allow the comparison between different ML approaches, it is essential to have common benchmarks and standard datasets available, similar to the open dataset MNIST that is often used in computer vision.
In order to effectively learn, ML algorithms will require a considerable amount of data. Furthermore, preferably standardized data generation/collection procedures should be created to allow reproducing the data. 
Research attempts in this direction include  \cite{o2016radio, rajendran2018electrosense}, showing that  synthetic generation of RF signals is possible, however some wireless problems may require to inhibit specifics of a real system in the data (e.g. RF device fingerprinting).

Therefore, standardizing these datasets and benchmarks remains an open challenge. Significant research efforts need to be put in building large-scale datasets and sharing them with the wireless research community.

\subsubsection{Standard Problems}
Future research initiatives should identify a set of common problems in wireless networks
to facilitate researchers in benchmarking and comparing their supervised/unsupervised learning algorithms.
These problems should be supported with standard datasets. For instance, in computer vision for 
benchmarking computer vision algorithms for image recognition tasks, the MNIST and ImageNet datasets are typically used. Examples of standard problems in wireless networks may be: wireless signal identification, beamforming, spectrum management, wireless network traffic demand prediction, etc.
Special research attention must be focused on designing these problems.

\subsubsection{Standard Data representation}
DL is increasingly used in wireless networks, however it is still unclear what the optimal data representation is. For instance, an I/Q sample may be represented as a single complex number, a tuple of real numbers or via the amplitude and phase values of their polar coordinates.
It is a debate that there is no one-size-fits-all data representation solution for every learning problem
\cite{kulin2018end}.
The optimal data representation might depend among other factors on the DL architecture, the learning objective and choice of the loss function \cite{o2017introduction}.

\change{\subsubsection{Standard evaluation metrics}
	After identifying standard datasets and problems, future research initiatives should identify a set of standard metrics for evaluating and comparing different ML models. 
	For instance, a set of standard metrics may be determined per standardized problem. Examples of standardized metrics might be: confusion matrix, F-score, precision, recall, accuracy, mean squared error, etc.
	In addition, the evaluation part may take into account other evaluation metrics such as: model complexity, memory overhead, training time, prediction time, required data size, etc.}

\subsection{Implementation of Machine Learning models in practical wireless platforms/systems}
There is no doubt that ML will play a prominent role in the evolution of future wireless networks.
However, although ML is powerful, it may be a burden when running on a single device. Furthermore, DL which has shown great success, requires significant amount of data to perform well, which poses extra challenges on the wireless network. 
It is therefore of paramount importance to advance our understanding of how to simply and efficiently integrate ML/DL breakthroughs within constrained computing platforms.
A second question that requires particular attention is which requirements does the network need to meet to
support collection and transfer of large volumes of data?

\subsubsection{Constraint wireless devices}
Wireless nodes, such as for instance seen in the IoT (e.g. phones, watches and embedded sensors), are typically inexpensive devices with scarce resources: limited storage resource, energy, computational capability and communication bandwidth. 
These device constraints bring several challenges when it comes to implement and run complex ML models.
Certainly, ML models with a large number of neurons, layers and parameters will necessarily require additional hardware and energy consumption not just for performing training but also for inference.

\paragraph{Reducing complexity of Machine Learning models}
ML/DL is well on its way to becoming mainstream on constraint devices \cite{lane2017squeezing}.
Promising early results are appearing across many domains, including hardware \cite{han2017ese}, systems and learning algorithms. For example, in \cite{mcdanel2017embedded} binary deep architectures are proposed, that are composed solely of 1-bit weights instead of 32-bit or 16-bit parameters, allowing for smaller models and less expensive computations. However, their ability to generalize and perform well in real-world problems is still an open question. 

\paragraph{Distributed Machine Learning implementation}
Another approach to address this challenge, may be to distribute the ML computation load across multiple nodes.
Some questions that need to be addressed here are: "Which part of the learning algorithms can be decomposed and distributed?", "How are the input data and output calculation results communicated among the devices?", "Which device is responsible for the assembly for the final prediction results?", etc.

\subsubsection{Infrastructure for data collection and transfer}
The tremendously increasing number of wireless devices and their traffic demands, require a scalable networking architecture to support large scale wireless transmissions.
The transmission of large volumes of data is a challenging task due to the following reasons: i) there are no standards/protocols that can efficiently deliver $>100T$ bits of data per second, ii) it is
extremely difficult to monitor the network in real-time, due to the huge traffic density in short time.

A promising direction in addressing this challenge is the concept of \textit{fog computing/analytics} \cite{mohammadi2018deep}. The idea of fog computing is to bring computing and analytics closer to the end-devices, which may improve the overall network performance by reducing or completely avoiding the transmission of large amounts of raw data to the cloud. Still, special efforts need to be devoted to employ these concepts in practical systems. \change{Finally, \textit{cloud computing} technologies (using virtualized resources, parallel processing and scalable data storage) may help reduce the computational cost when it comes to processing and analysis of data.}

\subsection{Machine Learning model accuracy in practical wireless systems}
Machine learning has been commonly used in static contexts, when the model speed is usually not a concern.
For example, consider recognizing images in computer vision. Whilst, images are considered as stationary data, wireless data (e.g. signals) are inherently time-varying and stochastic.
Training a robust ML model on wireless data that generalizes well is a challenging task due to the fact that wireless networks are inherently dynamic environments with changing channel conditions, user traffic demands and changing operating parameters (e.g. due to changes in standardization bodies). 
Considering that stability is one of the main requirements of wireless communication systems, 
rigorous theoretical studies are essential to ensure ML based approaches always work well in practical systems.
The open question here is "How to efficiently train a ML model that generalizes well to unseen data in such dynamically changing system?".
The following paragraphs discuss promising directions in addressing this challenge.

\subsubsection{Transfer learning}
With typical supervised learning a learned model is applicable for a specific scenario and likely biased to the training dataset. For instance, a learned model for recognizing a set of wireless technologies is trained to recognize only those technologies and also tight to the specific wireless environment characteristic where the data is collected. What if new technologies need to be identified? What if the conditions in the wireless environment change? Obviously, the ability of generalization of the trained learning models are still open questions.
How can we efficiently adapt our model to these new circumstances?

Traditional approaches may require to retrain the model based on new data (i.e. incorporating new technologies or specifics of a new environment together with new labels).
Fortunately, with the advances in ML it turns out that it is not necessary to fully retrain a ML model. A new popular method called \textit{transfer learning} may solve this. Transfer learning is a method that allows to \textit{transfer} the knowledge gained from one task to another similar task, and hereby alleviate the need to train ML models from scratch \cite{bacstuug2015transfer}. 
The advantage of this approach is that the learning process in new environments can be speeded up, with a smaller amount of data needed to train a good performing model.
In this way, wireless networking researchers may solve new but similar problems in a more efficient manner. For instance, if the new task requires to recognize new modulation formats, the model parameters for an already trained CNN model may be reused as the initialization for training the new CNN.

\subsubsection{Active learning}
Active learning is a subfield of ML that allows to update a learning model on-the-fly in a short period of time. For instance, in wireless networks, the benefit is that updating the model depending on the wireless networking conditions allows the model to be more accurate with respect to the current state \cite{yang2018active}.

The learning model adjusts its parameters whenever it receives new labeled data. The learning process stops when the system achieves the desired prediction accuracy. 

\subsubsection{Unsupervised/semi-supervised deep learning}
Typical supervised learning approaches, especially the recently popular deep learning techniques, require a large amount of training data with a set of corresponding labels. The disadvantage here is that so much data might either not always be available or comes at a great expense to prepare.
This is especially a time consuming task in wireless networks, where one has to wait for the occurrence of certain types of events (e.g. appearance of emission from a specific wireless technology or on a specific frequency band) for creating training instances to build robust models. At the same time, this process requires significant expert knowledge to construct labels, which is not a sufficiently automated process and generic for practical implementations.

To reduce the need for much domain knowledge and labeling data, \textit{deep unsupervised learning} \cite{rajendran2019unsupervised} and semi-supervised learning \cite{o2017semi} is recently used.
For instance, the AE (autoencoders) have become a powerful deep unsupervised learning tool \cite{o2016unsupervised}, which have also shown the ability to compress the input information by possibly learning a lower dimensional encoding of the input.
However, these new tools, require further research to fulfill their full
potentials in (practical) wireless networks.

\section{Conclusion}
\label{sec:concl}

\change{With the advances in hardware and computing power and the ability to collect, store and process massive amounts of data, machine learning (ML) has found its way into many different scientific fields, including wireless networks.
	The challenges wireless networks are faced with, pushed the wireless networking domain to seek more innovative solutions to ensure expected network performance. To address these challenges, ML is increasingly used in wireless networks.} 

\change{In parallel, a growing number of surveys and tutorials emerged on ML applied in wireless networks. We noticed that some of the existing works focus on addressing specific wireless networking tasks (e.g. wireless signal recognition), some on the usage of specific ML techniques (e.g. deep learning techniques), while others on the aspects of a specific wireless environment (e.g. IoT, WSN, CRN, etc.) looking at broad application scenarios (e.g. localization, security, environmental monitoring, etc.). 
Therefore, we realized that none of the works elaborate ML for optimizing the performance of wireless networks, which is critically affected by the proliferation of wireless devices, networks, technologies and increased user traffic demands.
We further noticed that some works are missing out the fundamentals, necessary for the reader to understand ML and data-driven research in general.
To fill this gap, this paper presented i) a well-structured starting point for non-machine learning experts, providing fundamentals on ML in an accessible manner, and ii) a systematic and comprehensive survey on ML for performance improvements of wireless networks looking at various perspectives of the network protocol stack. To the best of our knowledge, this is the first survey that comprehensively reviews the latest research efforts (up until and including 2019) in applying prediction-based ML techniques focused on improving the performance of wireless networks, while looking at all protocol layers: PHY, MAC and network layer. The surveyed research works are categorized into: radio analysis, MAC analysis and network prediction approaches. 
We reviewed works in various wireless networks including IoT, WSN, cellular networks and CRNs.
Within \textit{radio analysis} approaches we identified the following: automatic modulation recognition, and wireless interference identification (i.e. technology recognition, signal identification and emitter identification). \textit{MAC analysis} approaches are divided into: MAC identification, wireless interference identification and spectrum prediction tasks. \textit{Network prediction} approaches are classified into: performance prediction, and traffic prediction approaches.}

\change{Finally, open challenges and exciting research directions in this field are elaborated.
We discussed where standardization efforts are required, including standard: datasets, problems, data representations and evaluations metrics. Further, we discussed the open challenges when implementing machine learning models in practical wireless systems. Herewith, we discussed future directions at two levels: i) implementing ML on constraint wireless devices (via reducing complexity of ML models or distributed implementation of ML models) and ii) adapting the infrastructure for massive data collection and transfer (via edge analytics and cloud computing). Finally, we discussed open challenges and  future directions on the generalization of ML models in practical wireless environments.
}

We hope that this article will become a source of inspiration and guide for researchers and practitioners interested in applying machine learning for complex problems related to improving the performance of wireless networks.


\bibliography{mybibfile}

\end{document}